\pdfoutput=1

\documentclass[11pt]{article}

\usepackage[preprint]{acl}

\usepackage{times}
\usepackage{latexsym}
\usepackage{comment}
\usepackage[T1]{fontenc}

\usepackage[utf8]{inputenc}

\usepackage{microtype}

\usepackage{inconsolata}

\usepackage{graphicx}

\usepackage{subcaption}
\usepackage{booktabs}
\usepackage{tabularx}
\usepackage{amsmath}
\usepackage{makecell}
\usepackage{adjustbox}
\usepackage{multicol}
\usepackage{stfloats}
\usepackage{placeins}

\title{One world, one opinion? The superstar effect in LLM responses}

\author{Sofie Goethals \\
  University of Antwerp, Belgium \\
  \texttt{sofie.goethals@uantwerpen.be} \\\And
  Lauren Rhue \\
  Robert H. Smith School of Business, USA\\
}

\begin{document}
\maketitle
\begin{abstract}
As large language models (LLMs) are shaping the way information is shared and accessed online, their opinions have the potential to influence a wide audience. This study examines who the LLMs view as the most prominent figures across various fields, using prompts in ten different languages to explore the influence of linguistic diversity. Our findings reveal low diversity in responses, with a small number of figures dominating recognition across languages (also known as the "\textit{superstar effect}"). 
These results highlight the risk of narrowing global knowledge representation when LLMs retrieve subjective information.
\end{abstract}

\section{Introduction}
Large Language Models (LLMs) are becoming increasingly integrated into various aspects of society. With applications such as educational tools, writing assistance, and content generation, they have considerable potential to shape people's opinions and decisions~\citep{vida2024decoding, buyl2024large}. A report from the World Bank estimates that since the launch of ChatGPT, LLMs and other generative AI (GenAI) have already woven themselves into the daily routines of approximately half a billion people worldwide~\citep{liu2024earth}, which illustrates their widespread potential influence.

Although many LLMs originate in the United States, these LLMs are increasingly able to converse in multiple languages. 
These models can be leveraged for tasks such as synthesizing information~\citep{evans2024largelanguagemodelsevaluators}, replacing human input in surveys~\citep{bisbee2023synthetic}, or performing general information retrieval~\citep{zhu2023large}. They are thus transforming the way information is accessed and transmitted online~\citep{burton2024large}. 
However, the use of LLMs for these tasks may have unintended consequences such as narrowing the diversity of perspectives  \citep{shumailov2024ai, padmakumar2024doeswritinglanguagemodels, pedreschi2024human}.
Cultural opinions, such as who are celebrated figures, are expected to differ by language due to the language barriers in cultural diffusion. However, LLMs may share common embeddings and similar training sources leading to lower variation in responses.  

This paper  specifically focuses on how LLMs answer opinion-based prompts about celebrated figures in general and across different professions. These questions, such as "\textit{Who is the greatest person}?", reveal aspirational figures for society and for specific professional fields. We explore whether LLMs provide different responses to such opinion-based questions when we vary the language of the prompt. 
These types of questions do not have an objectively correct answer but rely heavily on cultural knowledge and values. 
We might expect a model to name very different celebrated individuals based on the language of inquiry, reflecting unique cultural values or locally relevant knowledge.

Additionally, we investigate whether the responses from the LLMs exhibit the "superstar effect". This effect, observed in various domains, suggest that a small number of figures dominate recognition and admiration, and this effect emerges as an artifact of technology mediation \citep{elberse2006superstars}. 
We examine the superstar effect by assessing the frequency and novelty of the LLM-generated names. Do the LLMs' responses reflect a language-specific spectrum of contributors from different cultures, or do the responses suggest a tendency to focus on a narrow subset of well-known individuals?
In case of the latter, this could sideline regionally important figures, ultimately narrowing global knowledge over time.
This effect of cultural homogenization is also discussed in other research~\citep{bommasani2022picking,durmus2023towards, alkhamissi2024investigating}.

Lastly, we analyse how the field of the profession influences the results.
Some fields are more international in nature than others due to their inherent characteristics and the processes by which knowledge is shared and recognized. Science, for example, is characterized by contributions that transcend cultural and national boundaries. This is largely due to the universal nature of scientific methods and principles, as well as the international collaborations in modern scientific research, making the contributions less tied to specific local contexts and more universally recognized~\citep{leydesdorff2008international}.  
Landmark contributions, such as Einstein's theory of relativity or Newton's laws of motion, have global relevance, irrespective of cultural or linguistic boundaries. In contrast, the arts and politics are often deeply embedded in local culture, history, and societal values~\citep{benedict2019patterns}. Artistic works, such as literature, music, or visual art, frequently draw upon the specific traditions, languages, and experiences of their creators. 
Politics is inherently a contested and subjective domain, shaped by diverse perspectives, ideologies, and cultural contexts. What may be celebrated as visionary leadership in one context can be condemned as authoritarianism in another.
As a result, we anticipate stronger consensus in scientific fields and more diversity in areas like the arts or politics.

Surprisingly, our findings reveal a substantial degree of consensus across languages, with many of the same individuals appearing regardless of the language used. 
For example, in every language, the most returned person for prompts about `\textit{mathematician}' is Isaac Newton. For prompts about `\textit{political figure}' the responses are more diverse, but Gandhi is the most returned person for almost every language except for Russian and Chinese (Mao Zedong) and for Urdu and Bengali (Nelson Mandela). We consistently find this concentration of names, which we refer to as the "superstar effect". For every profession, there is a single individual (or a small group of individuals) who appear in over two-thirds of the responses  across languages, LLMs and prompt variations. 
This illustrates that there is strong convergence in LLM outputs regardless of linguistic or model-specific differences. However, we did find some variation depending on the field of the profession, where science related professions lead to more consensus and professions related to art and politics to less. Our findings also indicate that languages with greater lexical similarity yield more aligned responses, suggesting a form of cultural consensus in the long tail of responses. 
We discuss potential causes and implications for this in Section~\ref{sec:discussion}.

The paper is structured as follows. We discuss related work in Section~\ref{sec:background}, and give more details about the materials, methods and metrics we use in Section~\ref{sec:m&m}. Our results are presented in Section~\ref{sec:results}. We discuss the implications and potential future research directions in Section~\ref{sec:discussion}, and end with listing the limitations of our study in Section~\ref{sec: limitations}.

 \section{Background} \label{sec:background}
There have been many studies that focus on how accurate LLMs are for multilingual input~\citep{watts2024pariksha}. As much of the initial training data of LLMs is written in English, LLMs tend to perform worse on non-English languages, particularly in under-resourced languages~\citep{ahuja2023mega,ahuja2023megaverse}. \citet{rajaratnamm2024nature} makes the analogy with a library predominantly filled with English books: a reader looking for resources in another language may struggle to find what they need—and LLMs face similar challenges.
We also investigate how LLM outputs vary across multilingual inputs, but we focus on alignment in opinions across languages rather than performance across languages, as there is no ground truth for these opinion-based tasks.

Another related area of research focuses on the cultural undertones in LLMs. One research stream evaluates language models' retention of culture-related commonsense by testing their responses to geographically diverse facts ~\citep{nguyen2023extracting,yin2022geomlama,keleg2023dlama}. Several studies investigate the cultural values that LLMs exhibit and find that these are more closely aligned with Western, Rich and Industrialized ideologies~\citep{cao2023assessing,tao2024cultural,  buyl2024large, rao2023ethical}.
\citet{vida2024decoding} highlight that the language of the prompt significantly influences LLM response behaviors, while \citet{alkhamissi2024investigating} demonstrate stronger cultural alignment when LLMs are prompted in the dominant language of a given culture.
 Additionally, \citet{durmus2023towards} compare LLM outputs to opinions from different countries on global issues. These studies study alignment in cultural values (often based on the World Values Survey~\citep{haerpfer2020world}) . More in line with our research is \citet{naous2023having} who find that when operating in Arabic,  LLM's exhibit a bias towards Western entities, failing in appropriate cultural adaptation.

This study explores LLMs' opinions about high achievers across different aspects of society because celebrities reflect the values of the society \cite{gorin2011celebrity, allison2016hero} and, under certain circumstances, can influence social norms \cite{cohen2024normative}. These notable figures, heroes with elevated social stature, are a means to represent cultural values in a way that is easy to communicate to all members of society and reflect the behaviors that should be modeled \cite{Sun2024heroes}. The identification of specific figures as the pinnacle of their field indicate the attributes that are valued in that field, and provide a lens to understand how others in this field are judged. 
Several studies in other fields observe the emergence of the "superstar effect" in the technology-mediated sales \citep{weeds2012superstars, brynjolfsson2010research}, where there is a concentration of demand among a few items and a very long tail among the others. 
This study will assess whether generative AI reveals similar trends when responding to opinion-based questions regarding notable figures.

\section{Materials and Methods} \label{sec:m&m}
\subsection{Materials}
This study uses three of the most well-known large language models, namely GPT-4 from OpenAI~\citep{achiam2023gpt}, Claude-3-Opus from Anthropic~\citep{anthropic2024claude}, and Llama-3.1-70B-Instruct from Meta~\citep{dubey2024llama}. To avoid any selection bias, we choose the ten most-used languages according to Wikipedia.~\footnote{\url{https://en.wikipedia.org/wiki/List_of_languages_by_total_number_of_speakers}}
We ask each language model in each of the chosen languages the following prompt: "\textit{Who is the \{adjective\} \{profession\}?}". The prompts cover fifteen professions with five descriptive adjectives. Profession is taken very broad here, and encompasses occupations such as writer or poet or more vague terms such as person or leader.  The used adjectives, professions and languages are shown in Table~\ref{tab:used_input}. 

\begin{table}[ht]
    \centering
    \small 
    \begin{tabularx}{\linewidth}{l l X}
        \toprule
        Languages & Adjectives & Professions \\
        \midrule
        English    & Greatest           & Leader \\
        Spanish    & Most Influential   & Military Leader \\
        Russian    & Most Important     & Poet \\
        Chinese    & Most Famous        & Philosopher \\
        Hindi      & Most Impactful     & Artist \\
        Arabic     &                    & Political Figure \\
        French     &                    & Composer \\
        Bengali    &                    & Writer \\
        Portuguese &                    & Physicist \\
        Urdu       &                    & Chemist \\
                   &                    & Economist \\
                   &                    & Medical Researcher \\
                   &                    & Mathematician \\
                   &                    & Computer Scientist \\
                   &                    & Person \\
        \bottomrule
    \end{tabularx}
    \caption{Languages, Adjectives, and Professions}
    \label{tab:used_input}
\end{table}

As a proxy for the cultural similarity of a language pair,
 we use the Similarity Database of modern lexicons of ~\citet{bella2021database}.
When languages have higher lexical similarity, it means they share a larger number of words with similar forms and meanings. This similarity often arises because the languages have a common linguistic ancestry (e.g., Latin for Romance languages), have historically interacted closely, or have borrowed words from each other over time~\citep{hock2009language}. 

\subsection{Methods}
\begin{figure} [ht]
    \centering
    \includegraphics[width=0.99\linewidth]{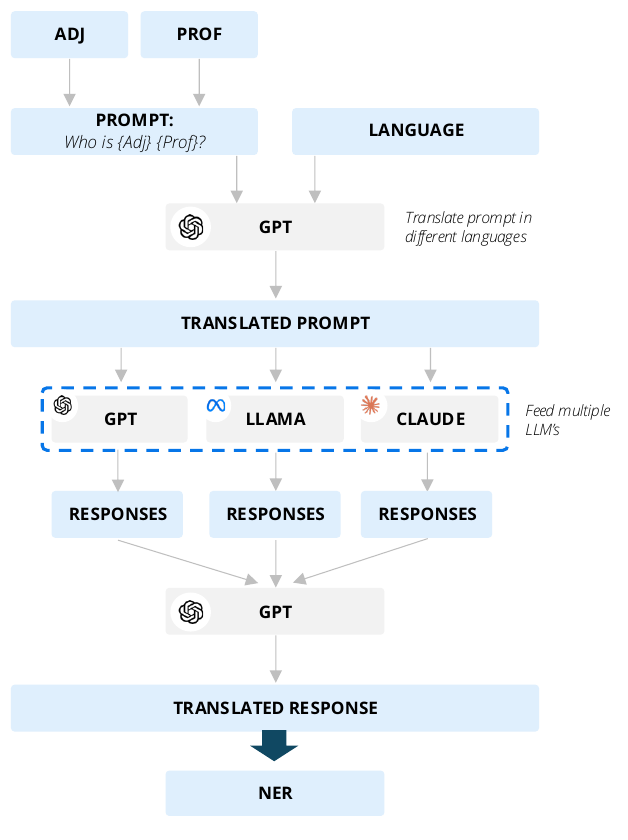}
    \caption{Used methodology}
    \label{fig:methodology}
\end{figure}

We present our methodological set-up in Figure~\ref{fig:methodology}. Each adjective and profession is combined into a prompt. For each prompt, we use GPT-4o to translate the initial prompt to the selected language. The translated prompt is submitted to each of the three LLMs and the LLMs' response is captured. Then, we use GPT-4o to translate the answer back to English. Based on the translated responses, we use Named Entity Recognition (NER) to identify the persons in the responses. We execute every combination of LLM, adjective, profession, and language five times, resulting in a total of 11,250 iterations.\footnote{Calculated as: 3 LLMs * 10 languages * 15 professions * 5 adjectives * 5 runs = 11, 250 iterations}

\paragraph{Large Language Models}
 This study prompts three popular LLMs, namely GPT-4 from OpenAI~\citep{achiam2023gpt}, Claude-3-Opus from Anthropic~\citep{anthropic2024claude}, and Llama-3.1-70B-Instruct from Meta~\citep{dubey2024llama}. 
 We use the default parameters for every LLM to reflect the way most users would use them.

\paragraph{Entity recognition}
To identify individuals mentioned in the responses, we apply Named Entity Recognition (NER) using the spaCy library ("\textit{en\_core\_web\_trf}"
 model).\footnote{ \url{https://spacy.io/api/entityrecognizer}} We process each translated response to extract named entities classified as `PERSON' labels.
 We perform manual verification to ensure different writing styles for one name refer to the same person.

\paragraph{Dimensionality reduction}
We use Multidimension Scaling (MDS) to generate the 2D-plots. MDS is a dimensionality reduction technique that projects high-dimensional data into a lower-dimensional space, preserving the pairwise distances between points as closely as possible~\citep{cox2000multidimensional}.

\subsection{Metrics} \label{sec:metrics}

We use \textbf{cosine similarity} to assess the similarity of the language representations of the LLMs' responses. Cosine similarity measures the angle between two vectors, where a smaller angle indicates greater similarity.  

We use the S\textbf{pearman correlation} coefficient~\citep{spearman1961proof} to measure the alignment between the lexical similarity and the average consensus between one language pair. This metric measures the strength and direction of a monotonic relationship between two variables by comparing their rank orders.

~\footnote{We use the implementation in \url{https://docs.scipy.org/doc/scipy/reference/generated/scipy.stats.spearmanr.html}.}

We measure the \textbf{novelty} of a set of responses $R$ as is done in the recommender literature~\citep{zhou2010solving,kaminskas2016diversity}:
\[
\text{Novelty}(R) = \frac{\sum_{i \in R} -\log_2 p(i)}{|R|}
\]
where $p(i)$ is the fraction of responses in the overall distribution that mention person $i$. For each name $i$ in the response set $R$, we will evaluate its novelty relative to the overall response distribution and subsequently compute the average novelty of the entire response set $R$.\footnote{In our experiments, we will measure this for every profession separately, and then take the average over the different professions.}

We use the \textbf{Gini coefficient}~\citep{dorfman1979formula} to measure the inequality in the distribution of name occurrences for each profession. This metric quantifies how unequal the distribution is by comparing the cumulative proportions of the population (which are all the unique persons that are returned for one profession) and the recognition they hold: 
\[
G = \frac{\sum_{i=1}^n \sum_{j=1}^n |x_i - x_j|}{2n^2 \mu}
\]
where:
\begin{itemize}
    \item \( n \) is the number of observations,
    \item \( x_i \) and \( x_j \) are the number of occurrences for individuals \( i \) and \( j \),
    \item \( \mu \) is the mean of the distribution.
\end{itemize}
A Gini coefficient of 0 reflects complete diversity in responses whereas values closer to 1 represent concentration (one person gets most of the recognition).

\section{Results} \label{sec:results}

\subsection{General results}
In this section, we discuss the effect of the prompt on the responses and present aggregate results. On average, 5.80 names are returned per prompt response. There are 2412 unique persons in total across all responses.

The details of each LLM are in tables in the Appendix.  The general results sorted by adjective and by LLM can be found in Table~\ref{tab:general_results_llmadjective}, by LLM and language in Table~\ref{tab:general_results_llmlanguage} and by LLM and profession in Table~\ref{tab:general_results_llmprofession}. The top 10 most occurring names for every profession can be found in Table~\ref{tab:results_names}.

\paragraph{How does the choice of LLM impact the responses?}

\begin{figure}[ht]
    \centering
    \includegraphics[width=0.99\linewidth]{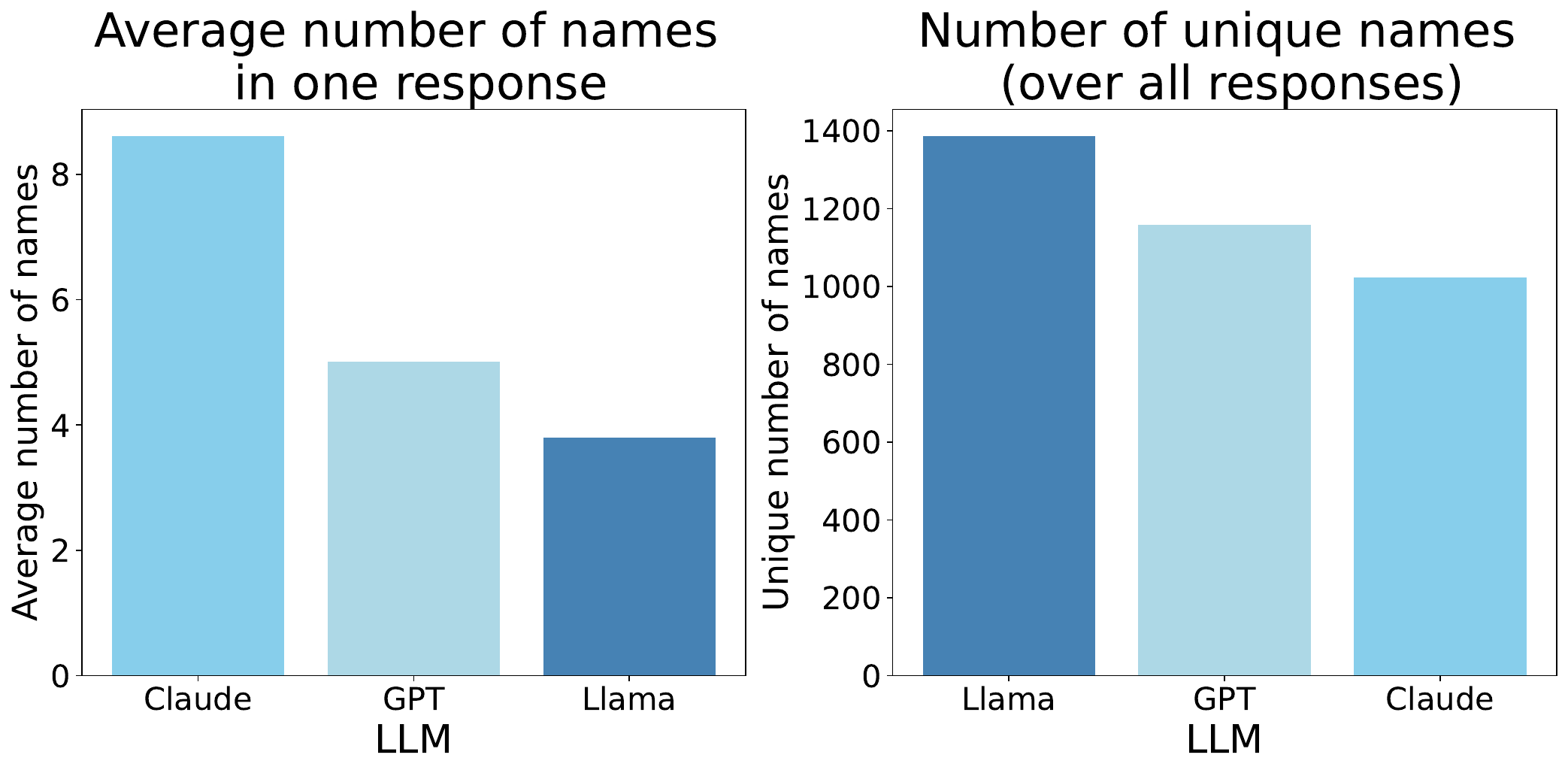}
    \vspace*{-8mm}
    \caption{General analysis by LLM}
    \label{fig:LLM_genanalysis}
\end{figure}

Figure~\ref{fig:LLM_genanalysis} demonstrates that the choice of LLM has a large impact on the results. On average, using Claude returns more than double the number of persons than using Llama, the LLM with the lowest number of responses. Despite returning the least persons on average for each run, Llama by far returns the most unique names across the different runs, adjectives and languages.

\begin{figure}[ht]
    \centering
    \begin{subfigure}{0.63\linewidth}
        \centering
    \includegraphics[width=\linewidth]{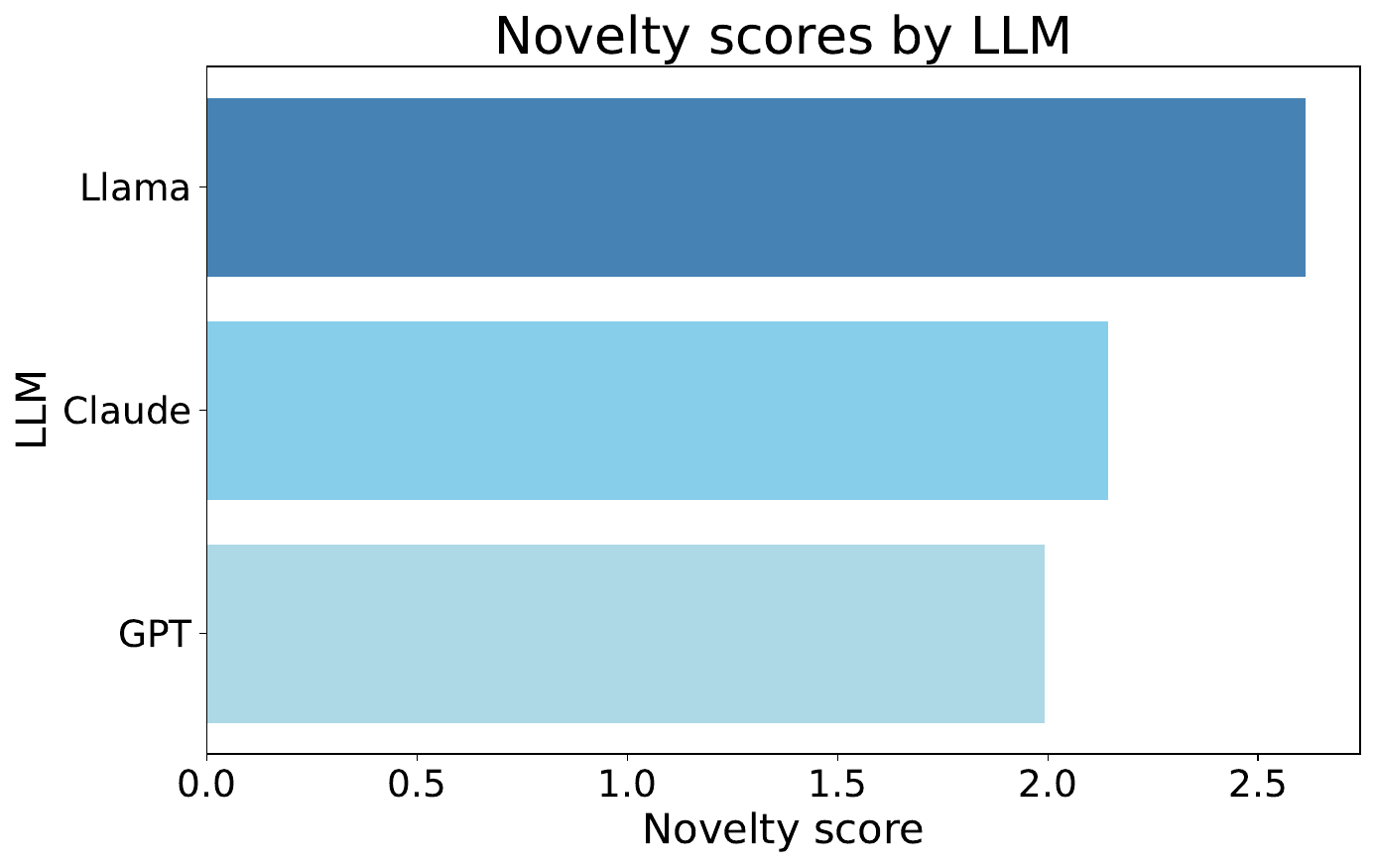}
        \caption{Novelty scores by LLM}
        \label{fig:novelty_llm_scores}
    \end{subfigure}
    \hfill
    \begin{subfigure}{0.35\linewidth}
        \centering
    \includegraphics[width=\linewidth]{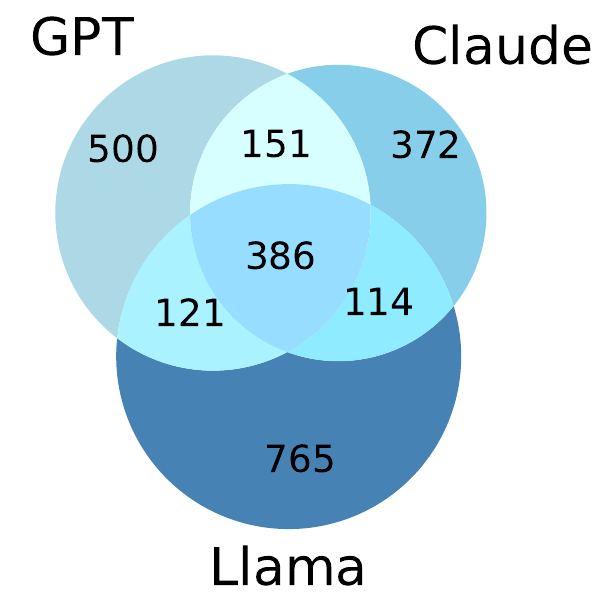}
        \caption{Overlap in names}
        \label{fig:novelty_llm_venn}
    \end{subfigure}
    \vspace*{-8mm}
    \caption{Novelty by LLMs}
    \label{fig:novelty_llm}
\end{figure}

In Figure~\ref{fig:novelty_llm}, we visualize the novelty of the LLM's responses in comparison to the overall response distribution for each profession. 
Llama clearly returns more names that are not present in the results of the other LLMs, which we can see both in the novelty scores in Figure~\ref{fig:novelty_llm_scores} and also in the Venn diagram in Figure~\ref{fig:novelty_llm_venn} that shows the overlap in unique names between the LLMs.

\paragraph{How do the adjectives impact the responses?}

\begin{figure} [ht]
    \centering
    \includegraphics[width=0.99\linewidth]{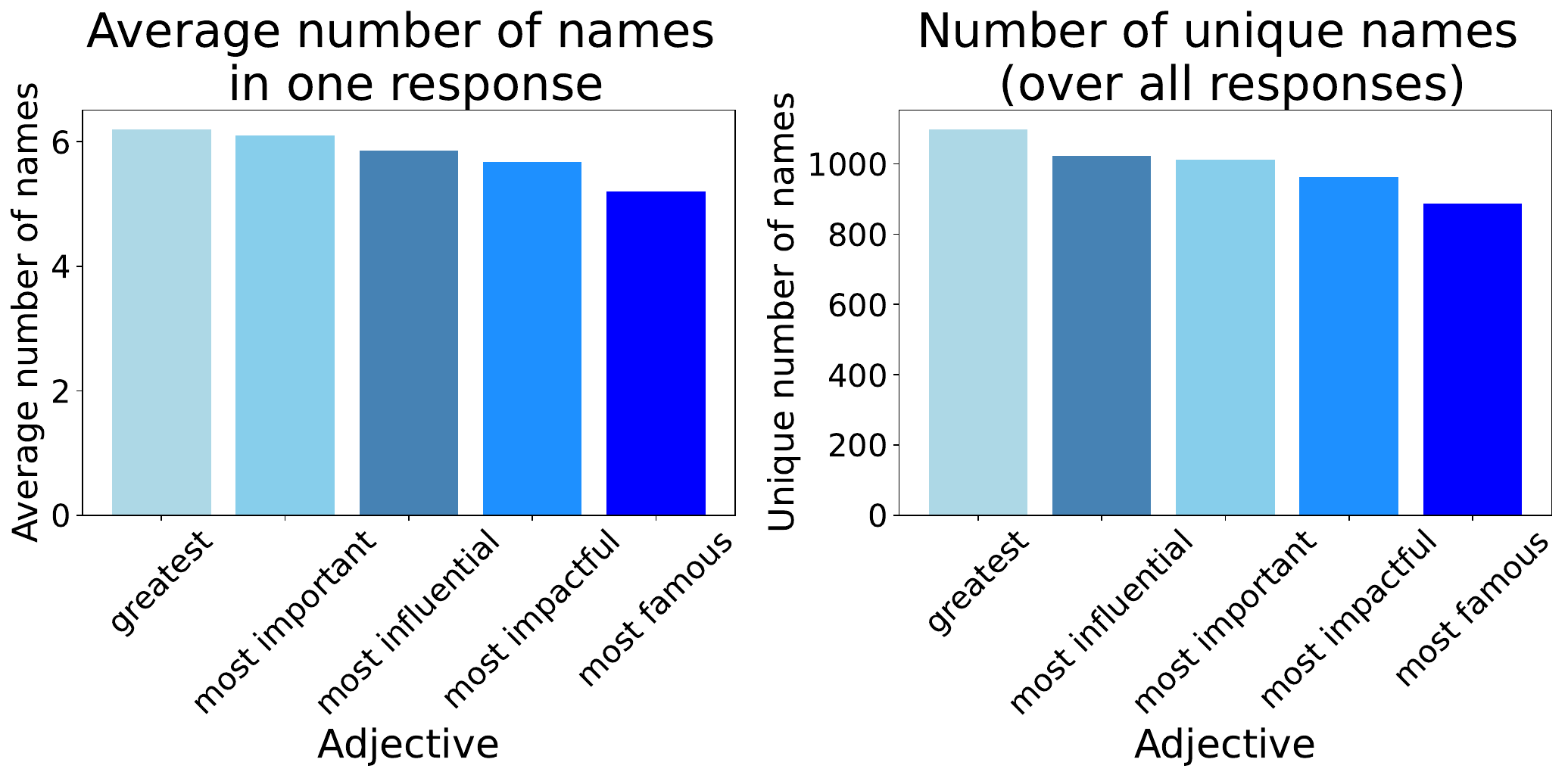}
    \vspace*{-8mm}
    \caption{General analysis by adjective}
    \label{fig:adjective_genanalysis}
\end{figure}

As explained in Section~\ref{sec:m&m}, we test different versions of the prompt by varying the adjective and running each version 5 times.  We see in Figure~\ref{fig:adjective_genanalysis} that the adjective `Greatest' leads to the most returned names on average, and that this is consistent across the LLMs (see Table~\ref{tab:general_results_llmadjective} in the Appendix).

The adjective `Most Famous' consistently returns the least names.
Similarly `Greatest' leads to the most unique names, and `Most Famous' to the least. We hypothesize that there may be more universal agreement on the criteria for fame whereas the criteria for adjectives like `Greatest' may be harder to define. 

\begin{figure}[ht]
    \centering
    \includegraphics[width=0.80\linewidth]{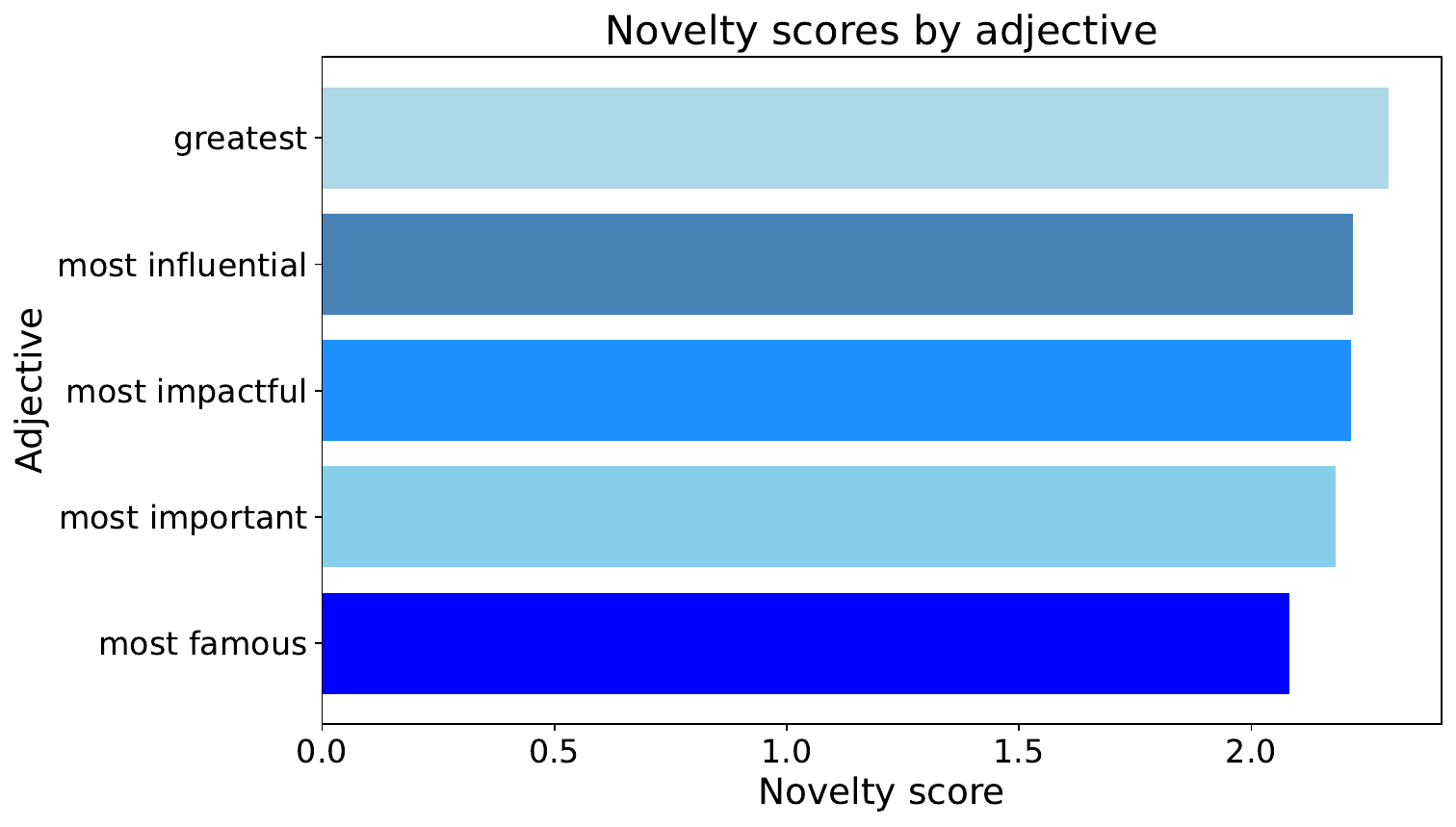}
    \vspace*{-2mm}
    \caption{Novelty of responses by adjective}
    \label{fig:adjective_novelty_scores}
\end{figure}

We see in Figure~\ref{fig:adjective_novelty_scores} that the adjective `Greatest' also leads to the most novel names across the professions, although there is only a slight difference. In summary, the adjectives yield slight differences in the responses but nothing extremely drastic. We also conduct a sentiment analysis to assess how the adjectives impact the polarity and subjectivity of the responses and present the results in Section~\ref{subsec:sentiment_analysis}.

\subsection{Prompt Language}

 \begin{figure}[ht]
    \centering
    \includegraphics[width=0.99\linewidth]{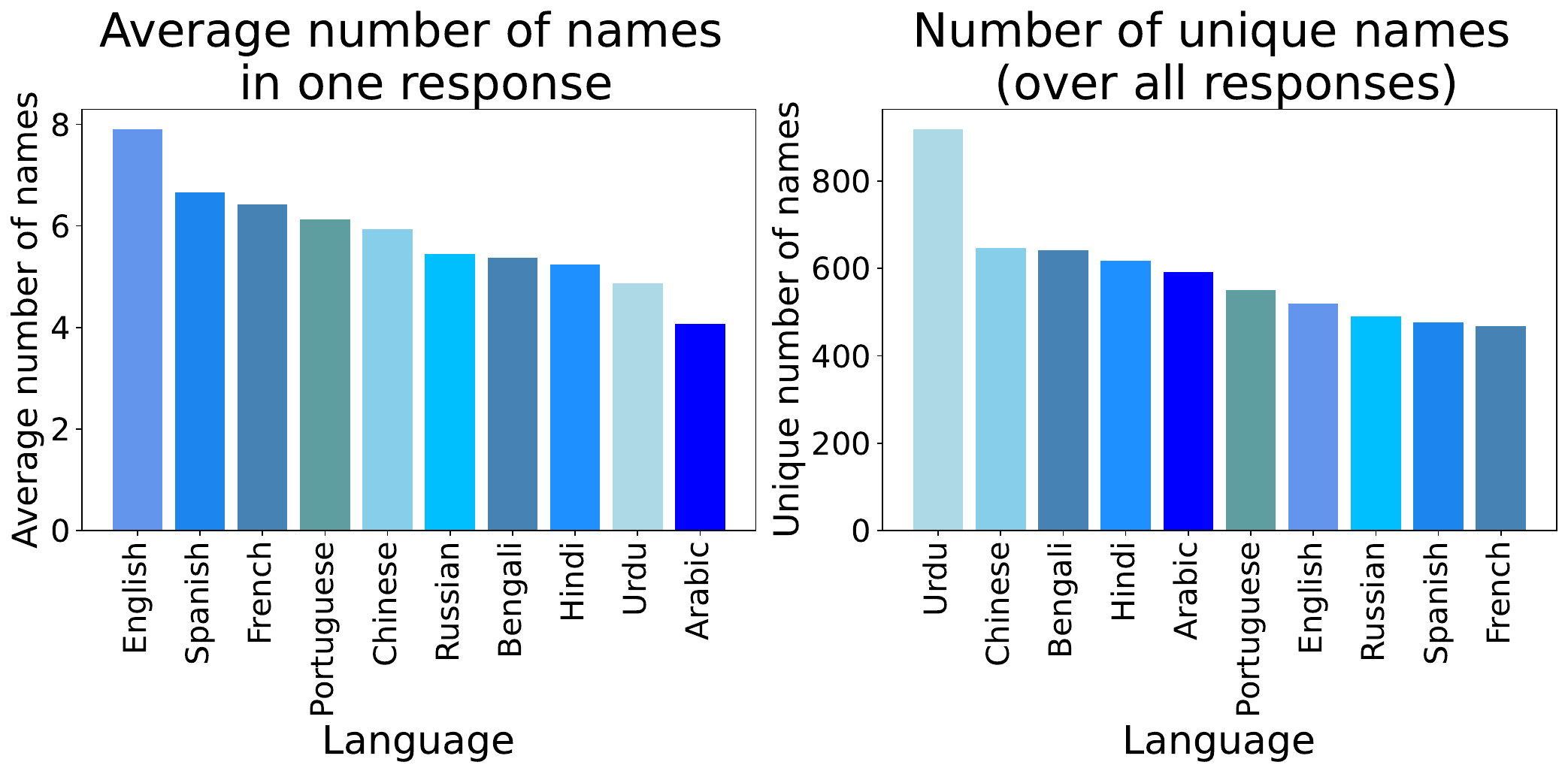}
    \vspace*{-8mm}
    \caption{General analysis by language}
    \label{fig:language_genanalysis}
\end{figure}

 Figure~\ref{fig:language_genanalysis} visualizes the general statistics for each language (aggregated across all LLMs). On average, we see that prompts in English return the most names and prompts in Arabic return the least, and this pattern holds true across LLMs.  Urdu returns more unique names compared to the other languages, a pattern that we also see for every LLM separately but that is most striking for Llama. (For details, see Table~\ref{tab:general_results_llmlanguage} in the Appendix).

\begin{figure} [ht]
    \centering
    \includegraphics[width=0.99\linewidth]{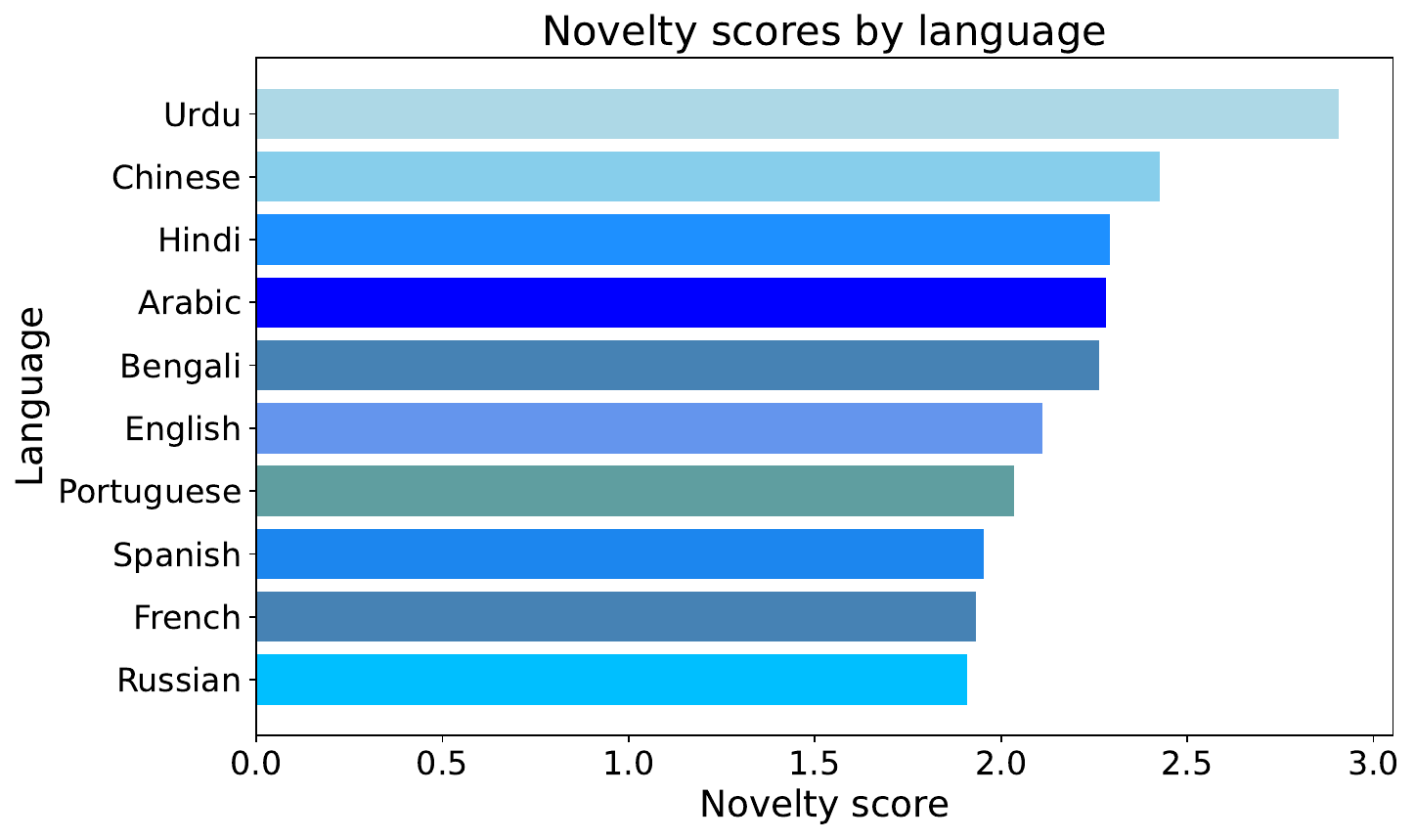}
    \vspace*{-8mm}
    \caption{Novelty of the responses by language} 
    \label{fig:language_originality_scores}
\end{figure}

We analyse the novelty of the results for one language compared to the results of all languages. 
 Figure~\ref{fig:language_originality_scores} displays the novelty scores of each language. We see that prompts in Urdu or Arabic tend to return more novel names than prompts in French or Spanish.

\paragraph{Which languages have similar responses?}
\begin{figure*}[hbt]
    \centering
    \begin{subfigure}{0.32\linewidth}
        \centering
    \includegraphics[width=\linewidth]{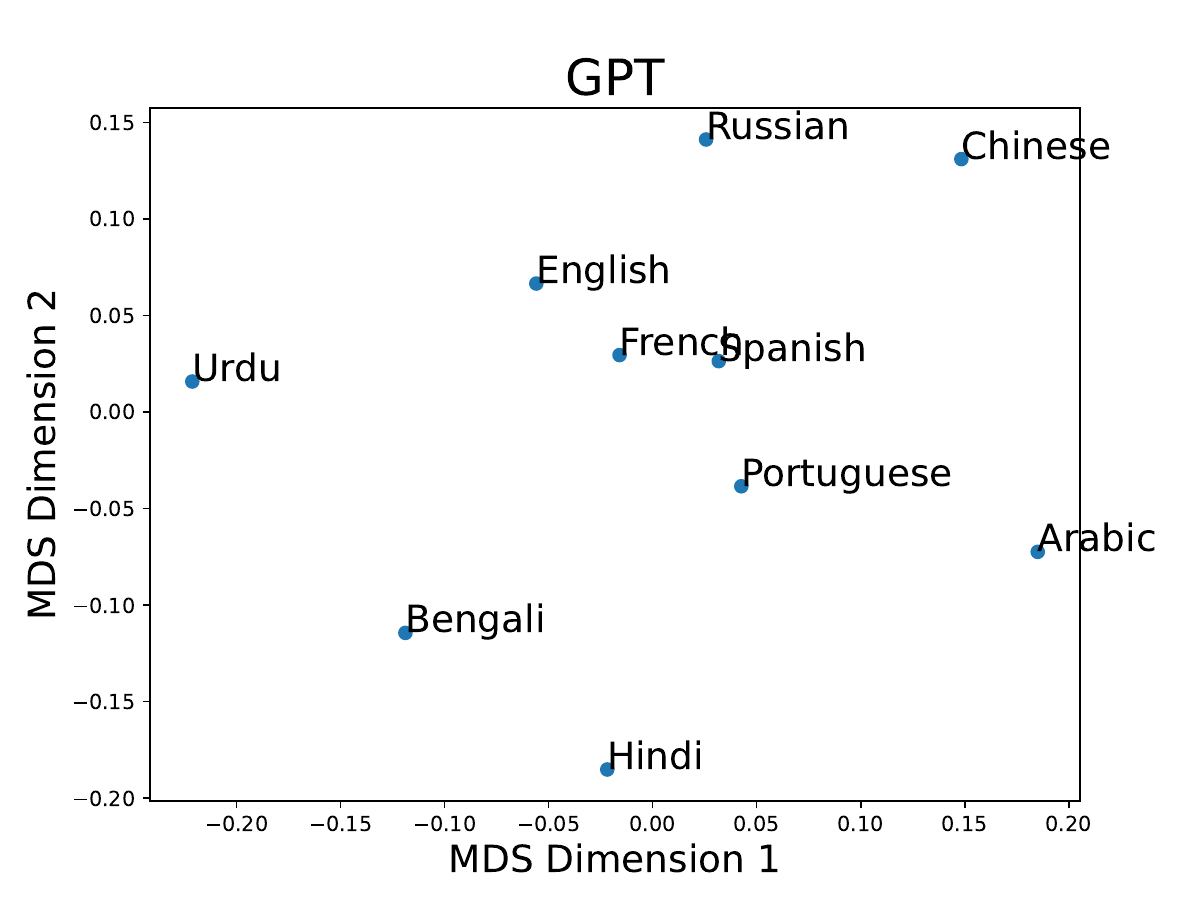}
        \caption{GPT}
        \label{fig:gpt-similarity-lang}
    \end{subfigure}
    \hfill
    \begin{subfigure}{0.32\linewidth}
        \centering
    \includegraphics[width=\linewidth]{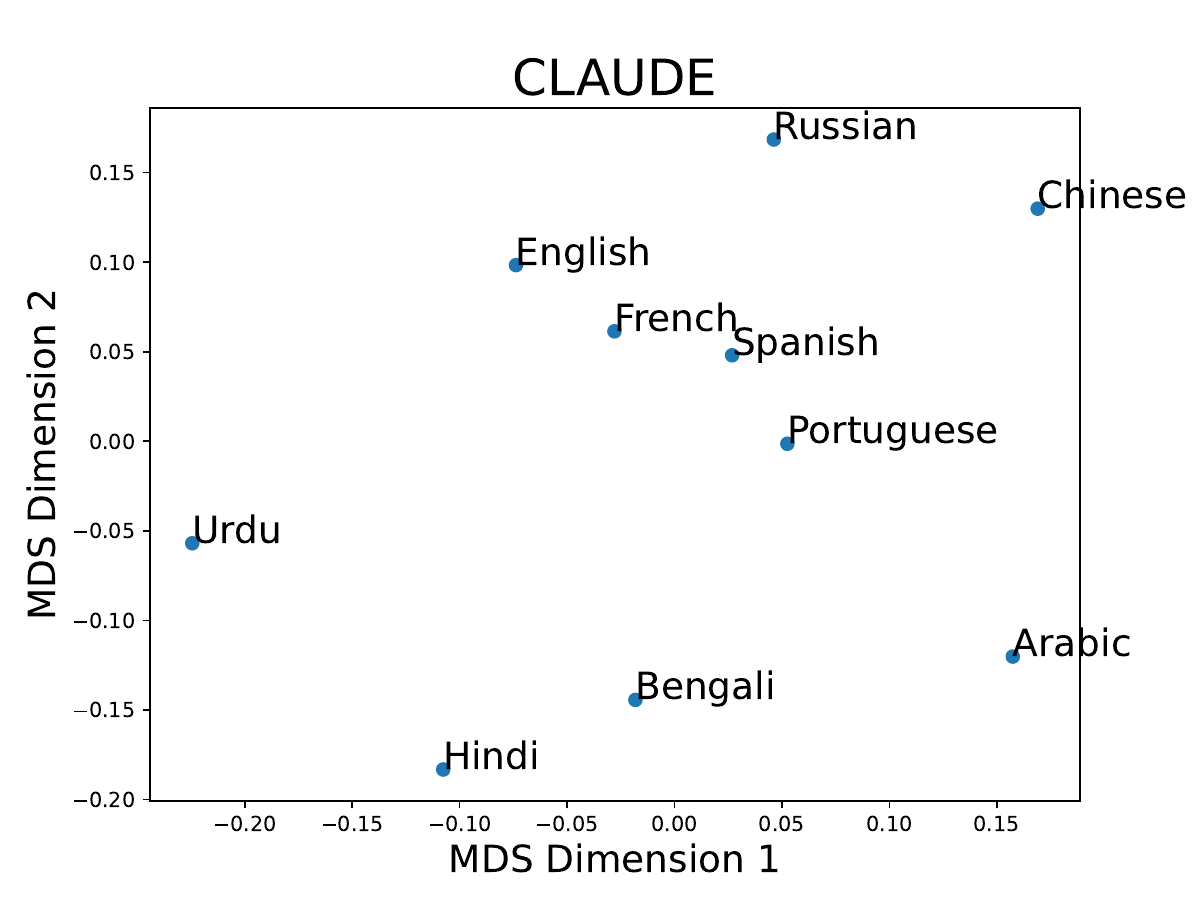}
        \caption{Claude}
        \label{fig:claude-similarity-lang}
    \end{subfigure}
    \hfill
    \begin{subfigure}{0.32\linewidth}
        \centering
    \includegraphics[width=\linewidth]{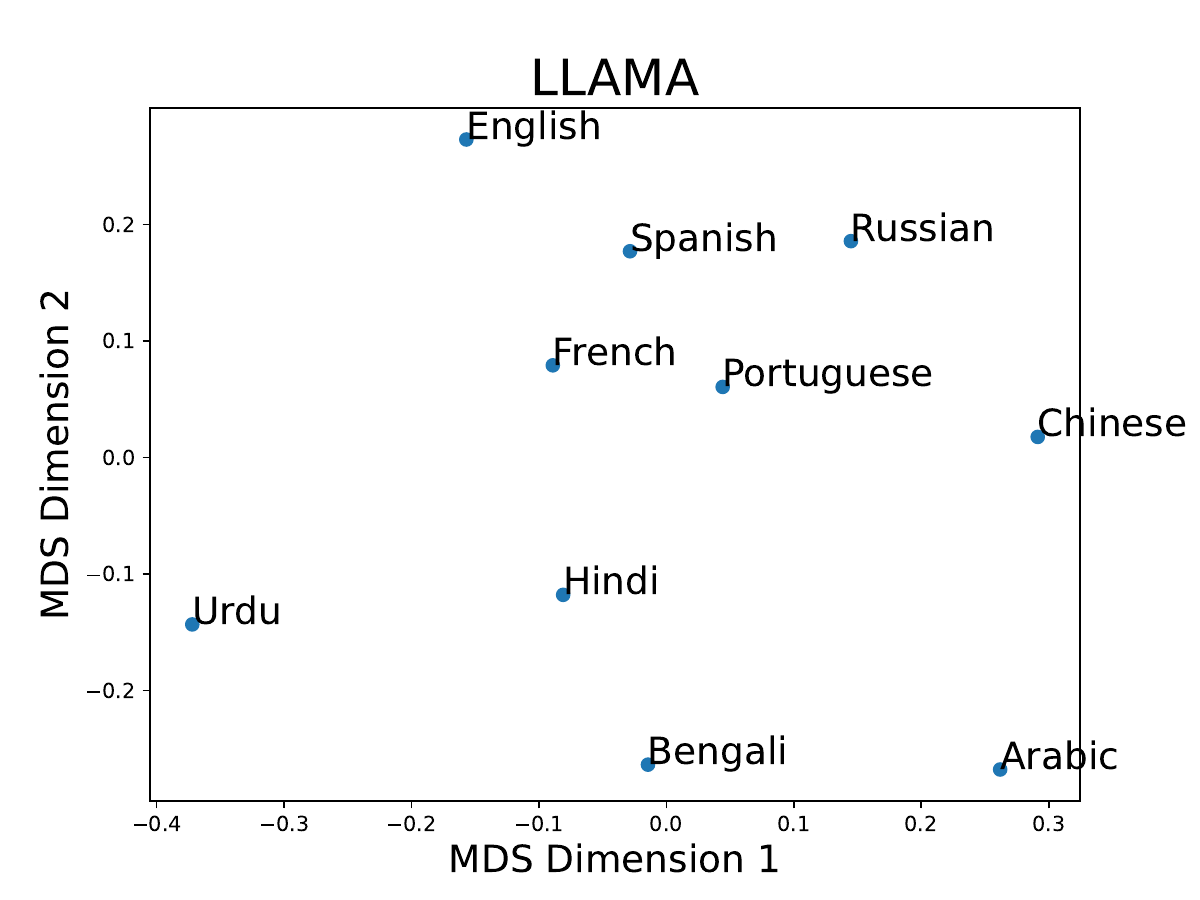}
        \caption{Llama}
        \label{fig:llama-similarity-lang}
    \end{subfigure}
    \vspace*{-2mm}
    \caption{Similarity of the responses by language and LLM}
    \label{fig:language-similarity}
\end{figure*}

To understand which languages yield similar responses, we measure the consensus between all the languages with the cosine similarity metric and visualise the results with MDS in a 2D-plot for every LLM in Figure~\ref{fig:language-similarity}.  Responses from languages with similar cultures and history are closer together. For example, for each of the LLMs, the responses from languages from European origin appear in one centroid. The responses from Asian languages appear on the periphery, more distant from the European language responses in the center. 

To verify this pattern statistically, we compare these results with the Similarity Database of Modern Lexicons~\citep{bella2021database}.
 We verify for each LLM separately whether there is a correlation between the average consensus of each language pair, and how similar their lexicons are to each other.  

\begin{table}[ht]
    \centering
    \begin{tabular}{c|ccc}
        LLM & GPT & Claude & Llama  \\ \hline
        Correlation & 0.450 & 0.532 & 0.401\\
        p-value & 0.002** & 0.002**  &  0.006**
    \end{tabular}
    \caption{Spearman correlation between the similarity in modern lexicons and the consensus between languages}
    \label{tab:stat_analysis}
\end{table}
We see a significant correlation for every LLM in Table~\ref{tab:stat_analysis}.  This means that languages with higher lexicon similarity tend to have more consensus on which persons should be venerated.

\subsection{Profession}

The general results for each profession and LLM can be found in Figure~\ref{fig:profession_genanalysis}. The full details are in  Table~\ref{tab:general_results_llmprofession} in the Appendix. We divide the professions according to their overarching categories (Science, Politics, Art and General) to see how this influences their results. We compare the average response rate and see that general, vague `professions' such as `Person' lead to the most names per response, while science-related professions such as physicist or chemist consistently generate  fewer names in the returned responses. 

\begin{figure} [ht]
    \centering
    \includegraphics[width=0.99\linewidth]{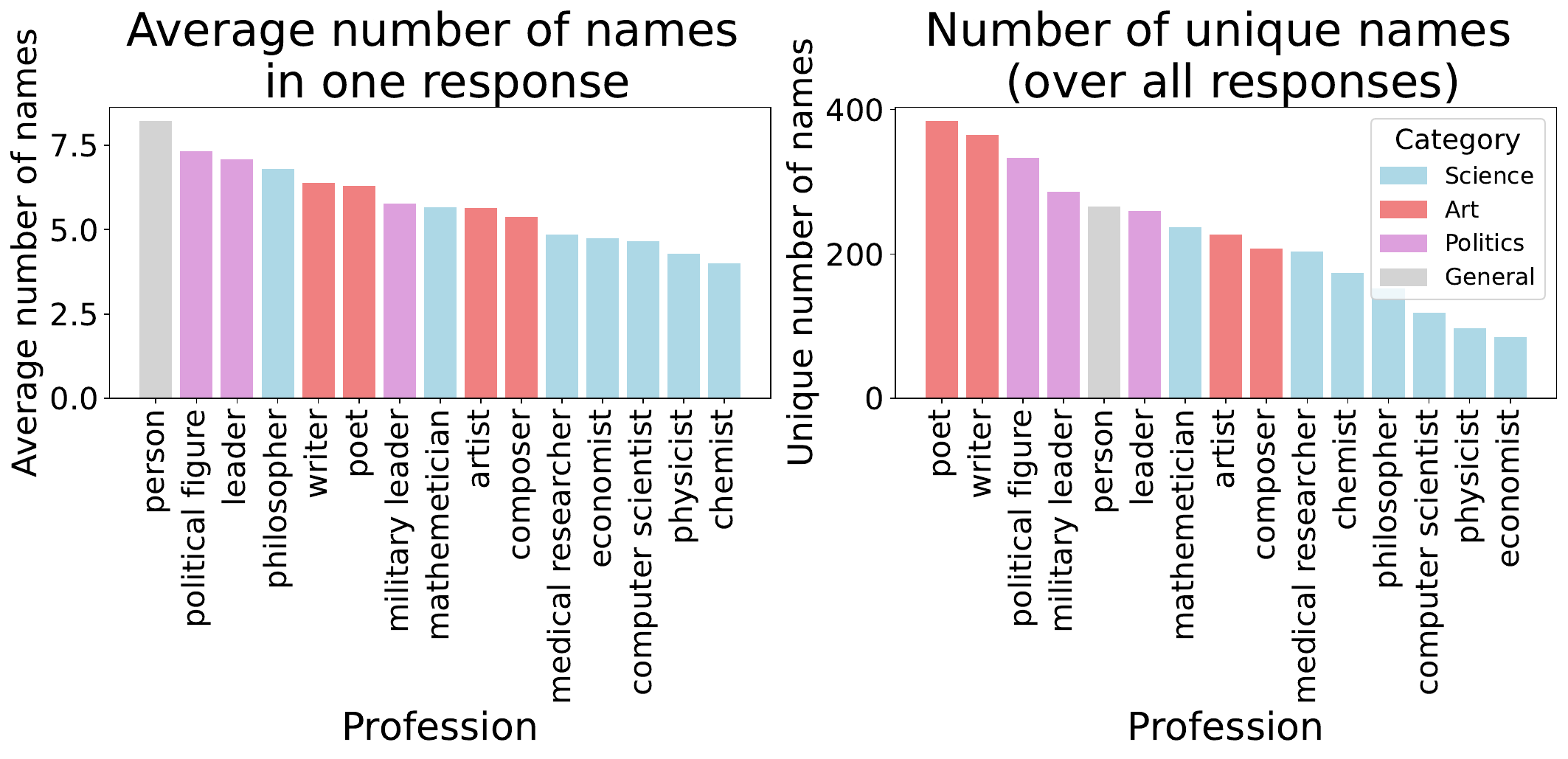}
    \vspace*{-8mm}    
    \caption{General analysis by profession and field}
    \label{fig:profession_genanalysis}
\end{figure}

\paragraph{Is there a trend in responses by professions?}
As anticipated, we note that accomplishments in some professions are more likely transcend national and linguistic boundaries whereas accomplishments in other professions are more linked to local cultures. 

\begin{figure} [ht]
    \centering
    \includegraphics[width=0.99\linewidth]{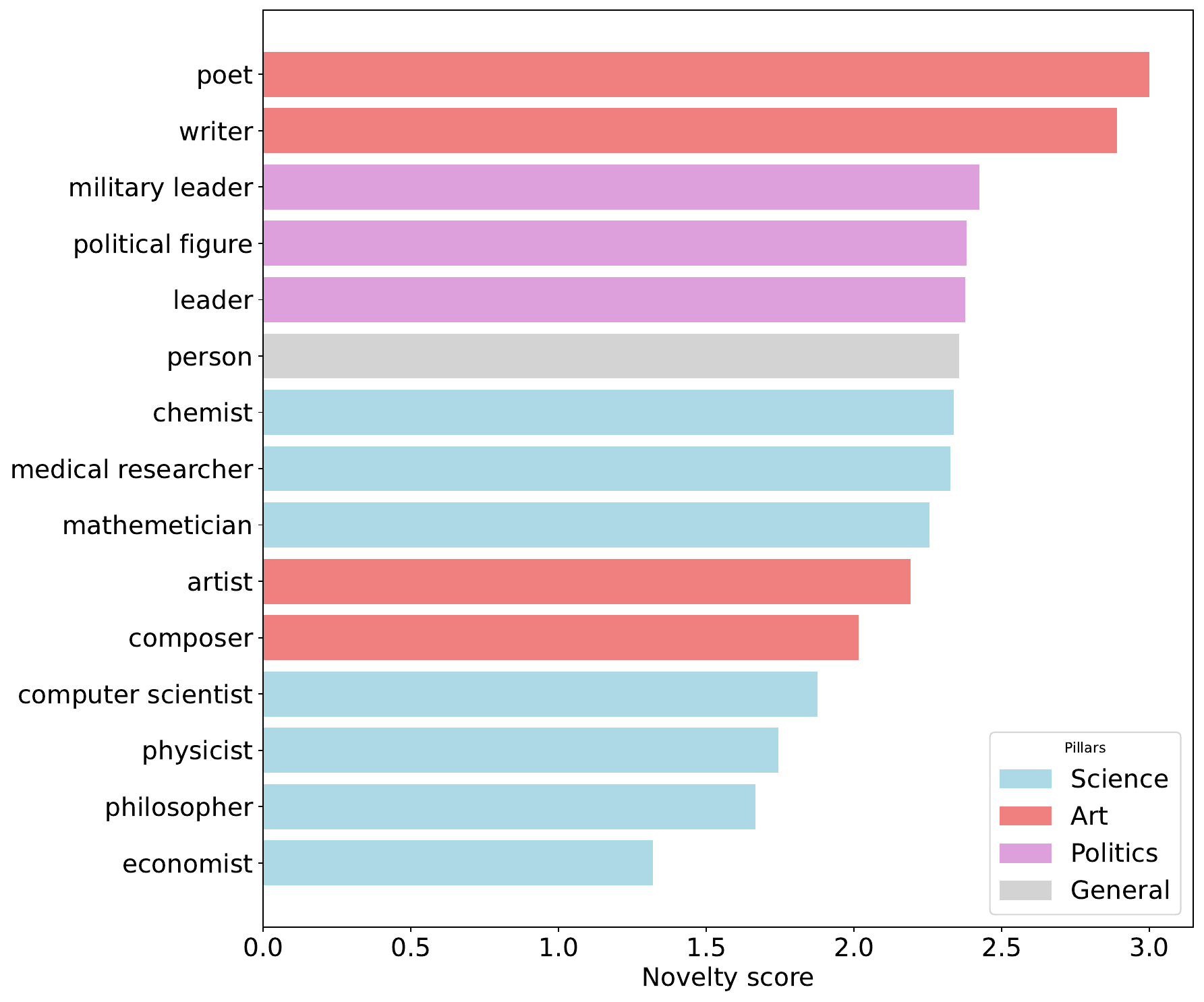}
    \vspace*{-8mm}
    \caption{Novelty by field and languages}
    \label{fig:originality_profession}
\end{figure}

 Figure~\ref{fig:profession_genanalysis} suggests that professions in the sciences tend to yield fewer unique names compared to fields such as politics or art. 
For each profession, we also calculate the average novelty score across the languages and visualize this in Figure~\ref{fig:originality_profession}.~\footnote{We calculate the novelty score for each language (for each profession compare the novelty of the response set of that language with the overall response set), and then calculate the novelty score of one profession as the average over the different languages.} We see that fields such as `Poet' and `Writer' that heavily depend on the language, and more subjective fields such as `Military leader' and `Political figure' lead to the most novel names across the languages.
This aligns with our expectation that science represents a field with more globally recognized contributors whose influence transcends national boundaries, whereas politics and art are fields that often reflect more localized and culturally specific perspectives.

\paragraph{Do we see the superstar effect for any of the professions?}

\begin{figure*}[ht]
    \centering
    \includegraphics[width=0.70\linewidth]{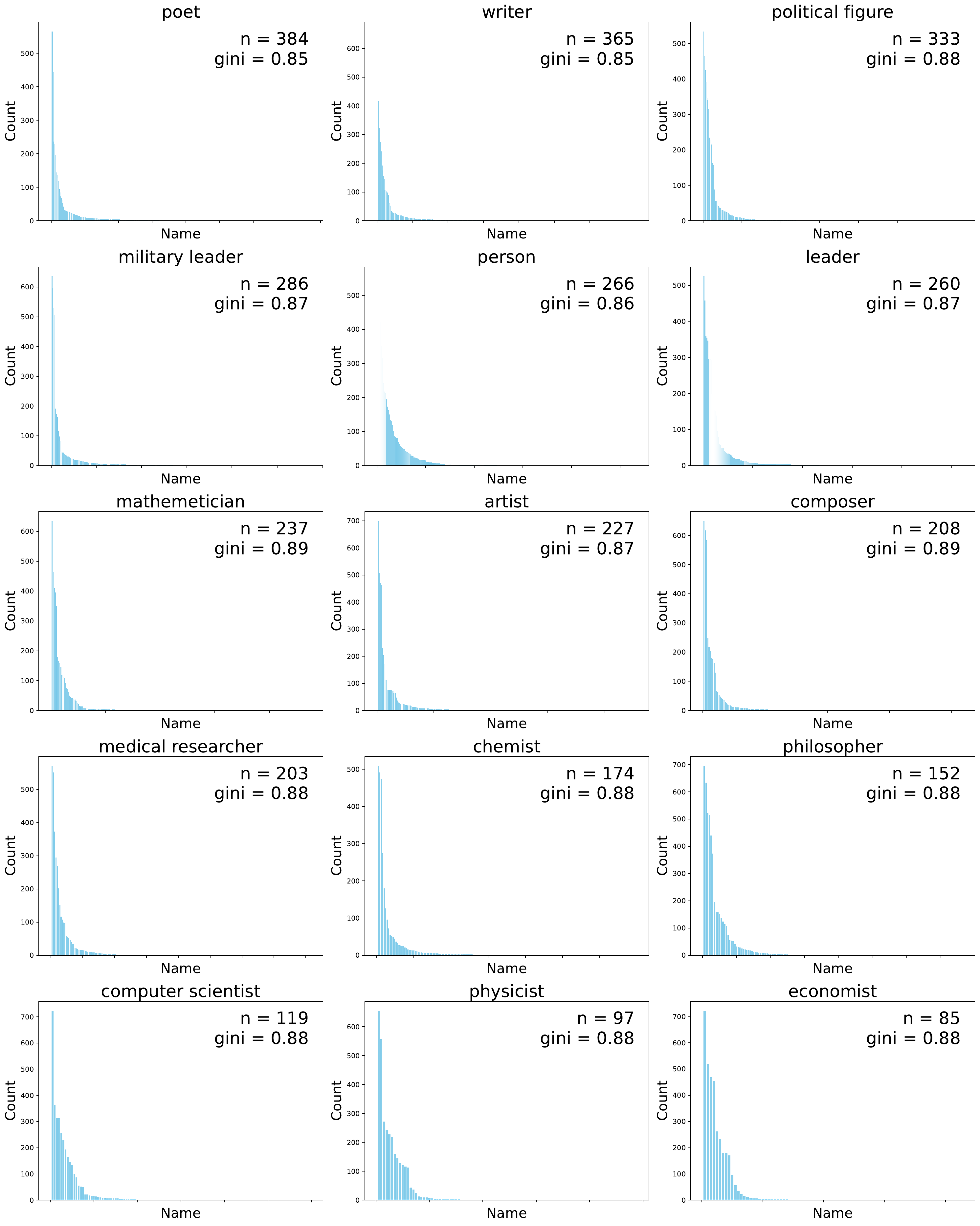}
    \caption{Frequency distribution for every profession (across LLMs). The number of unique names is depicted in the right corner ($n$).}
    \label{fig:freq_occ_combined}
\end{figure*}

We analyze the frequency distribution of names returned for each profession. 
The aggregate analysis for the LLMs combined can be found in Figure~\ref{fig:freq_occ_combined}. The results for each LLM separately can be found in the Appendix (Figures~\ref{fig:freq_occ_gpt} -~\ref{fig:freq_occ_llama}). For Figure~\ref{fig:freq_occ_combined}, a total of 750 responses are generated for every profession.\footnote{3 LLMs * 10 languages * 5 adjectives * 5 runs = 750 responses} All the distributions share a sharp peak and a long tail, indicating a few people who are consistently included in the answers across all parameters. 
 For instance, Alan Turing is present in $96.4\%$ ($n=723$) of the responses for computer scientist (so across different adjectives, runs, languages and LLMs), and Adam Smith is present in $96.3\%$ ($n=722$) of the responses for economist. For every profession, there is a person that is present in more than 2/3  of the responses ($n > 500$). (See Table~\ref{tab:results_names} in the Appendix for details.) 
 All professions exhibit a long tail of names that are returned only a few times or even just once. To quantify the concentration in the responses, we calculate the Gini coefficient for each profession and consistently find values higher than 0.70, which indicates very unequal distributions.~\footnote{The Gini values are displayed in Figures~\ref{fig:freq_occ_combined} -~\ref{fig:freq_occ_llama}.}

\section{Discussion} \label{sec:discussion}
In this study, we identify three important factors when using LLMs to generate opinions about prominent figures: the influence of lexical similarity, the presence of the superstar effect in LLM responses and the impact of the field of the profession. 
Our findings indicate that languages with greater lexical similarity tend to show higher consensus on which individuals are venerated. This outcome aligns with expectations, as linguistic overlap often reflects cultural interconnectedness. 
The superstar effect is particularly evident in global fields, such as computer science and physics. These fields often yield consensus on globally renowned figures, such as Alan Turing or Albert Einstein, who dominate the LLM responses. However, even in more locally appreciated professions, such as the arts, LLMs exhibit a preference towards 
dominant figures, often from the Western hemisphere. For example, William Shakespeare consistently emerges as the most celebrated writer in every language, overshadowing culturally specific authors.

This pattern highlights a broader trend: LLMs prioritize popular opinions, often at the expense of cultural diversity. Such behavior is consistent with the narrowing of knowledge discussed in prior literature.
 \citet{shumailov2024ai} illustrate the risk of homogeneity in AI-generated content, as when the AI predicts what to generate, the path of least resistance is an averaging of the content in its source material. Similarly, \citet{doshi2024generative} argue that while using AI can boost individual creativity, it comes at the expense of less varied content overall. 
 \citet{pedreschi2024human} warn that human-AI coevolution might lead to a loss of diversity in generated content, while \citet{burton2024large} discuss how the use of large language models can reshape collective intelligence by reducing functional diversity among individuals.
This type of knowledge homogeneity could stem from the training data and processes underlying these models. Training datasets may overrepresent globally influential figures or sources from a few dominant cultures.
Moreover, the architecture of LLMs promotes shared embeddings and parameters across languages, resulting in consistent output. Cross-linguistic transfer learning~\citep{lai2024llmsenglishscalingmultilingual} amplifies this effect by encoding general, cross-linguistic knowledge rather than language-specific nuances.

While this paper does not aim to prescribe whether LLMs should prioritize producing more consensus or embracing greater diversity in their opinions, it is crucial to consider some of its implications.
For example, teenagers writing a school paper about "a great writer" might no longer consult their parents or teachers but instead ask an LLM. If the models consistently suggest a narrow set of globally renowned authors like Shakespeare or Tolstay, it could limit exposure to regionally significant writers, leading to a narrowing of global knowledge over time. Different levels of consensus or diversity might be appropriate depending on the context.
For example, in fields like physics or mathematics, a higher degree of consensus might be desirable due to its universal nature, while in literature or politics, diversity and cultural specificity might be more suited.

With LLMs rapidly changing the way information is accessed and shared online, it is vital to proactively anticipate some of its unintended consequences.
This study raises questions about the balance between global consensus and cultural specificity in AI-generated content and encourages users to be aware of this behavior when relying on LLMs to retrieve information that can involve subjective perspectives.

\paragraph{Future research directions}
Our main direction of future research would be to compare these results with human responses. This would be done by conducting a survey with participants in each language, and comparing the diversity in opinions. Do people who speak these languages agree with the assessment of LLMs on who should be celebrated for their achievements in these fields? Do they produce more or less diverse opinions?

\section{Limitations} \label{sec: limitations}
As can be seen in our methodological set-up, all responses are translated back to English before the consequent analysis. The manner of translation could have some influence on the results. However, as we do not use the actual responses (except for the sentiment analysis in the Appendix) but only the named entities present in the response, the translation manner will have less impact. The fact that we only investigate the persons present in the response can also be seen as a limitation, as we do not analyse the remainder of the response or the ordering in which the persons occur.
\citet{naous2023having} also found that NER works better for Western persons than for Arabic persons, which could influence the returned persons from other cultures. 

The choice of languages and models is also a limiting factor. To avoid any selection bias, we opted for the ten most spoken languages, and three of the most popular LLMs. 
However, we could extend the analysis to some less popular languages as well. The lower language support may lead to an increase in the superstar effect, as there may be less local cultural awareness.
Similarly, an interesting follow-up experiment could be using LLMs developed in different countries and see how this would affect these results.

Besides this, we use the current version of the LLMs for our experiments, which presents challenges for reproducibility as they can be updated at anytime, potentially altering the results.
Lastly, we used the default parameters for every LLM but varying some of the parameters (such as temperature) could also influence the diversity of the response. We opted for the default parameters to reflect the way that most users would interact with the LLMs.

\section*{Acknowledgments}
This research was funded by Flemish Research Foundation (grant number 1247125N). 
\newpage

\bibliography{custom}
\clearpage
\newpage
\appendix

\section{Appendix}
\label{sec:appendix}

\subsection{Sentiment analysis} \label{subsec:sentiment_analysis}
We use TextBlob to analyze the sentiment of the text responses, measuring polarity (the positivity or negativity) and subjectivity (the degree of opinion versus fact) for each response~\citep{loria2018textblob}. The sentiment of the responses is analyzed after the LLMs' responses are translated back into English. In this case, we use the complete text responses and not only the returned names.

\begin{figure} [ht]
    \centering
     \begin{subfigure}{0.99\linewidth}
        \centering
    \includegraphics[width=\linewidth]{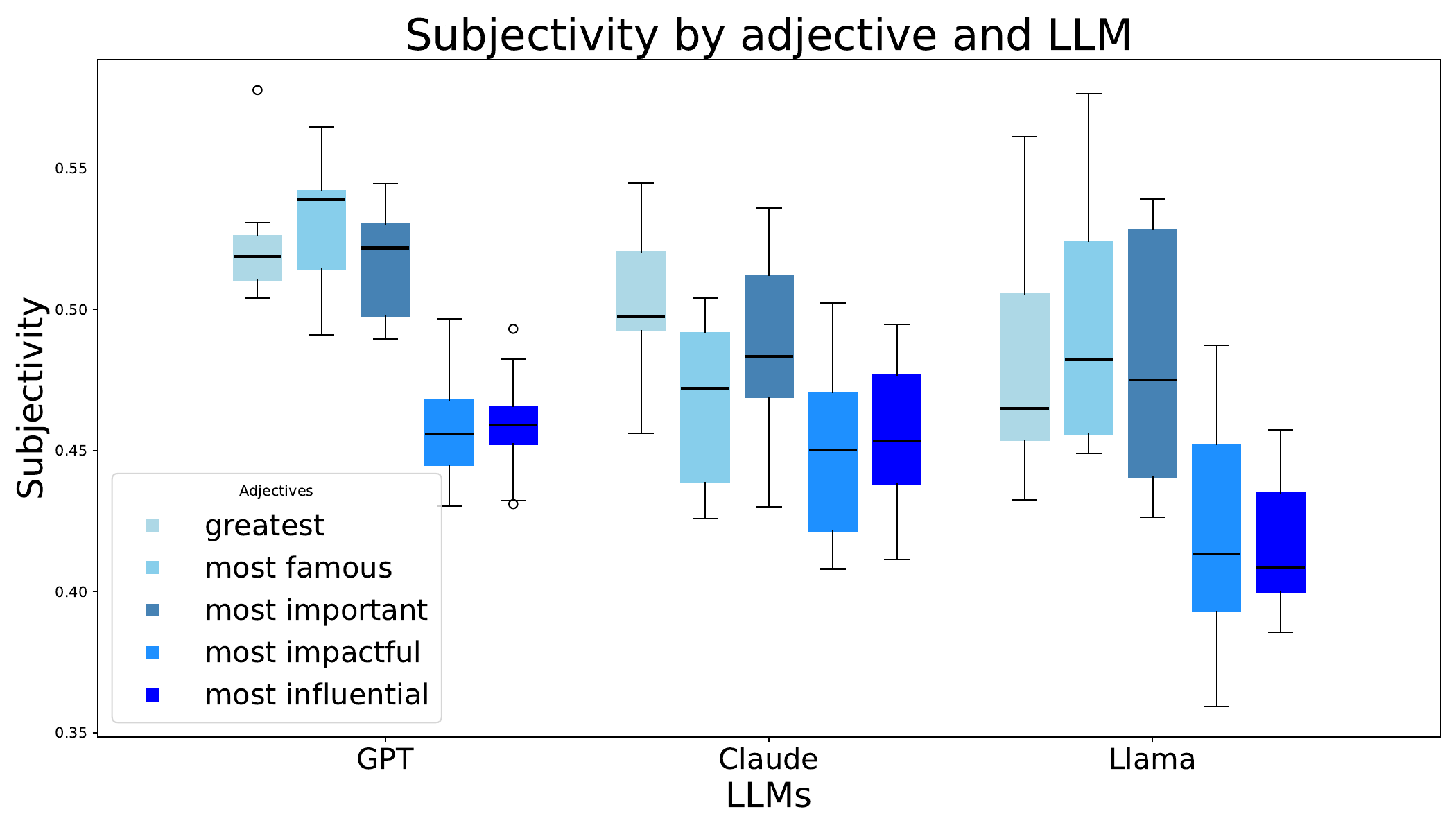}
    \end{subfigure}
    \hfill
    \begin{subfigure}{0.99\linewidth}
        \centering
    \includegraphics[width=\linewidth]{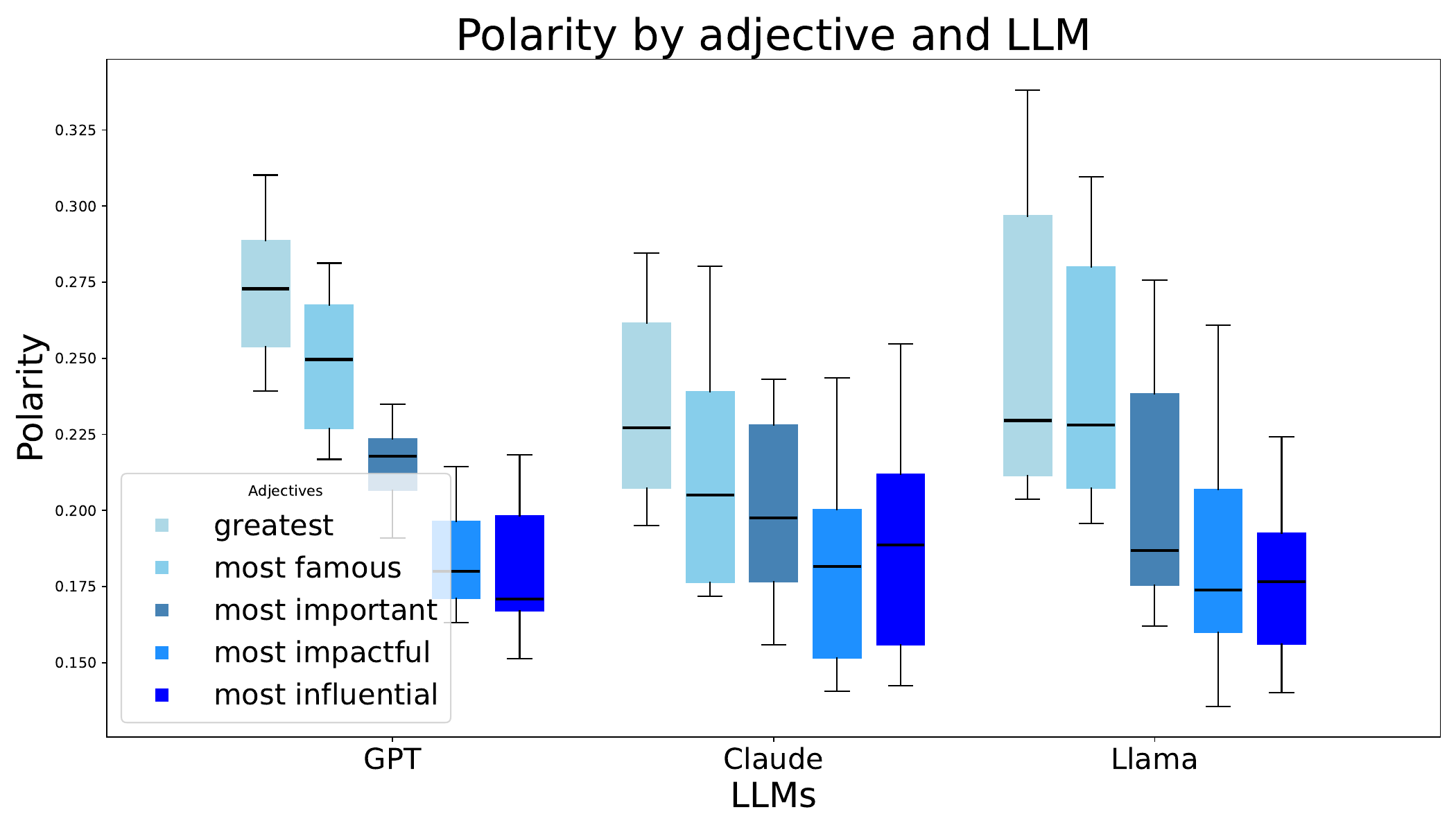}
    \end{subfigure}
    \hfill
    \caption{Sentiment analysis}
    \label{fig:sentiment}
\end{figure}

We see in Figure~\ref{fig:sentiment} that the average polarity and subjectivity of the text response can vary a lot depending on the adjective that was used in the prompt.
\newpage
\subsection{Additional results}
In this Section, we present some of the additional results. We display the full results by LLM and adjective in Table~\ref{tab:general_results_llmadjective}, by LLM and language in Table~\ref{tab:general_results_llmlanguage} and by LLM and profession in Table~\ref{tab:general_results_llmprofession}.

Table~\ref{tab:general_results_llmadjective} reveals that Claude generates the highest number of unique names within a single adjective. However, it produces the fewest unique names when considering results across different adjectives. This suggests that Claude's outputs are the least affected by variations in prompt phrasing (adjective choice).

Table~\ref{tab:general_results_llmlanguage}  illustrates that Claude produces the most unique names within one language, but the least unique names when we look at the results of all the languages combined. This suggests that the output of Claude is the least affected by the language of the prompt as well.

\begin{figure*}[ht]
    \centering
    \includegraphics[width=0.82\linewidth]{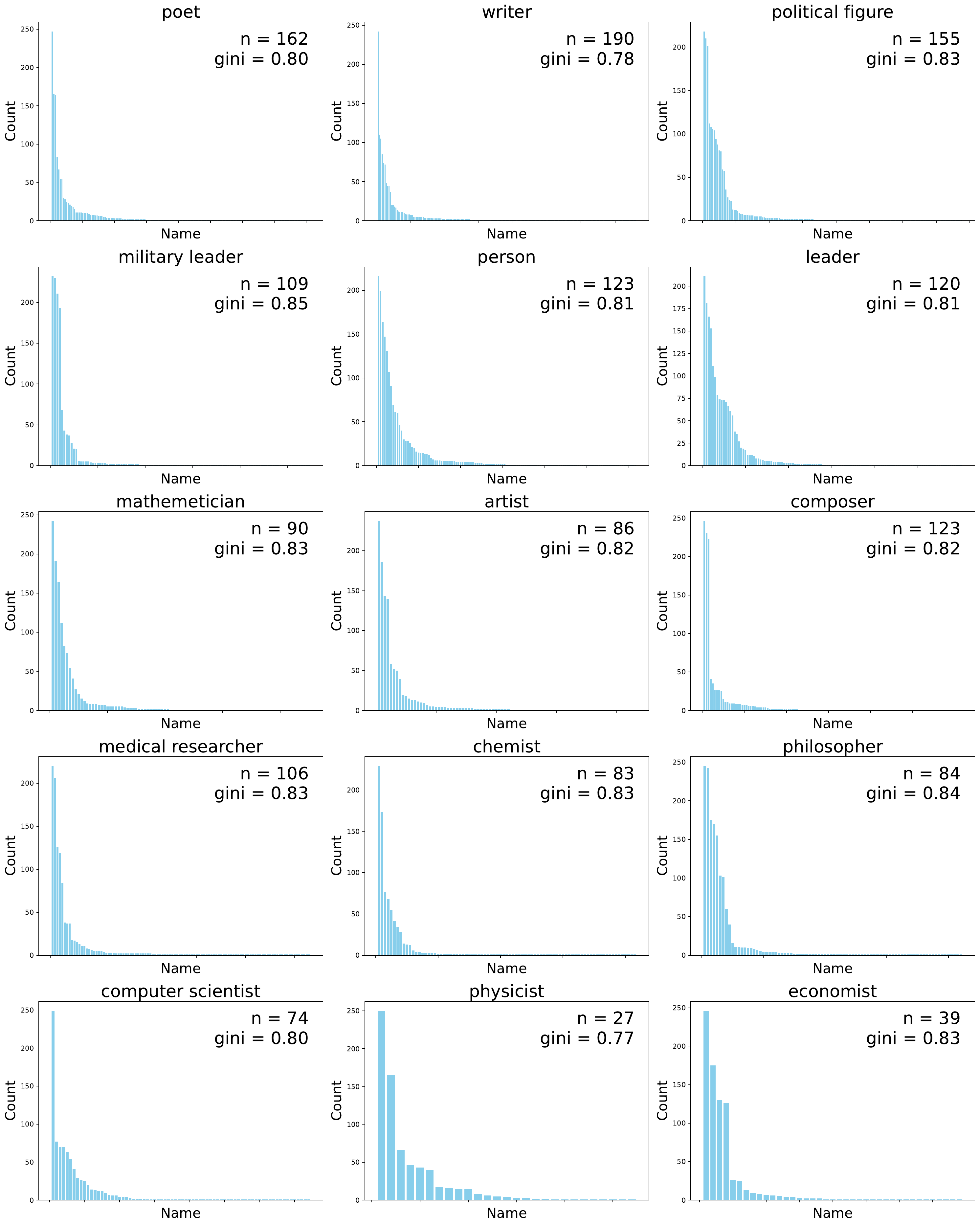}
    \caption{Frequency distribution for every profession (GPT)}
    \label{fig:freq_occ_gpt}
\end{figure*}

\begin{figure*}[ht]
    \centering
    \includegraphics[width=0.82\linewidth]{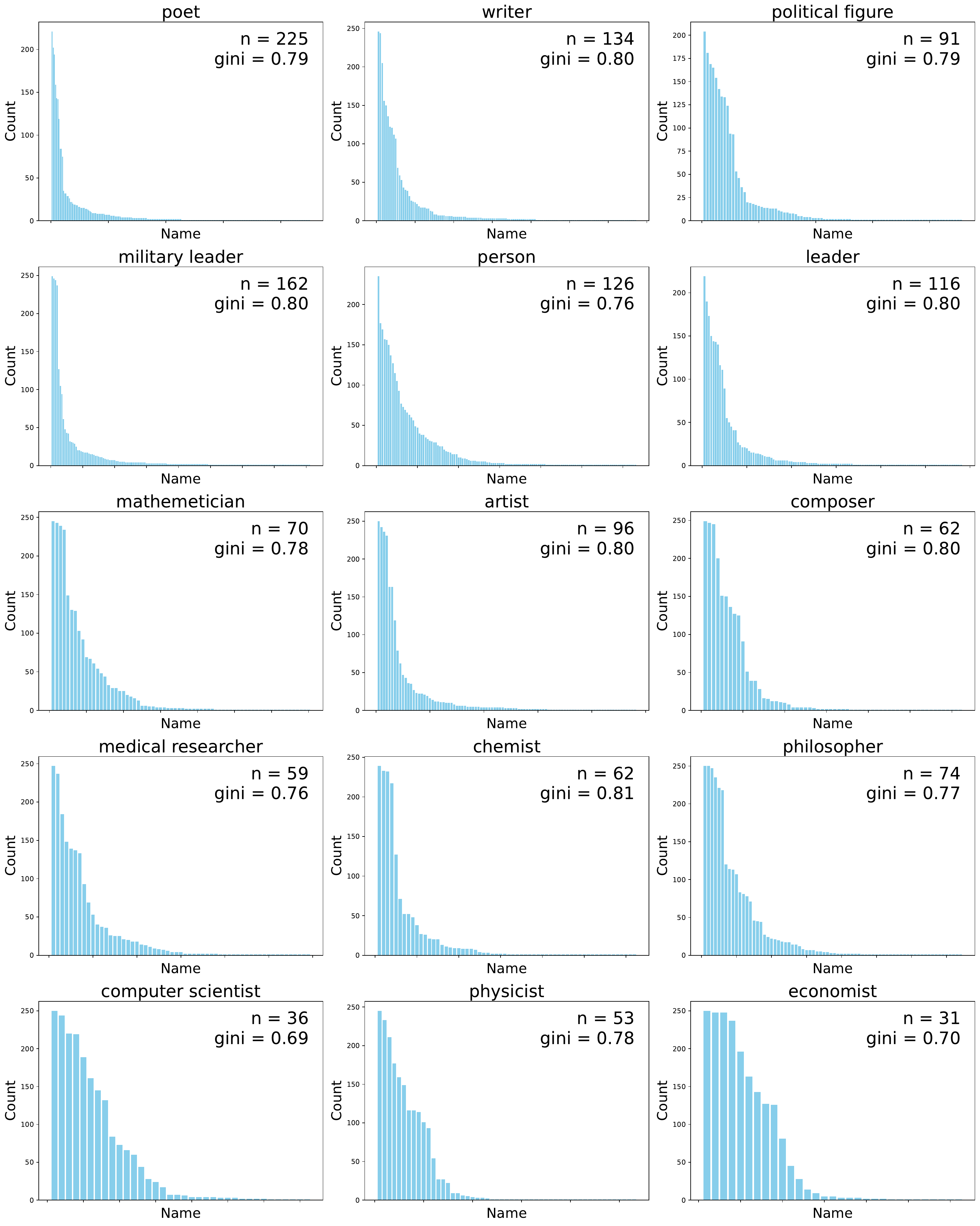}
    \caption{Frequency distribution for every profession (Claude)}
    \label{fig:freq_occ_claude}
\end{figure*}

\begin{figure*}[ht]
    \centering
    \includegraphics[width=0.82\linewidth]{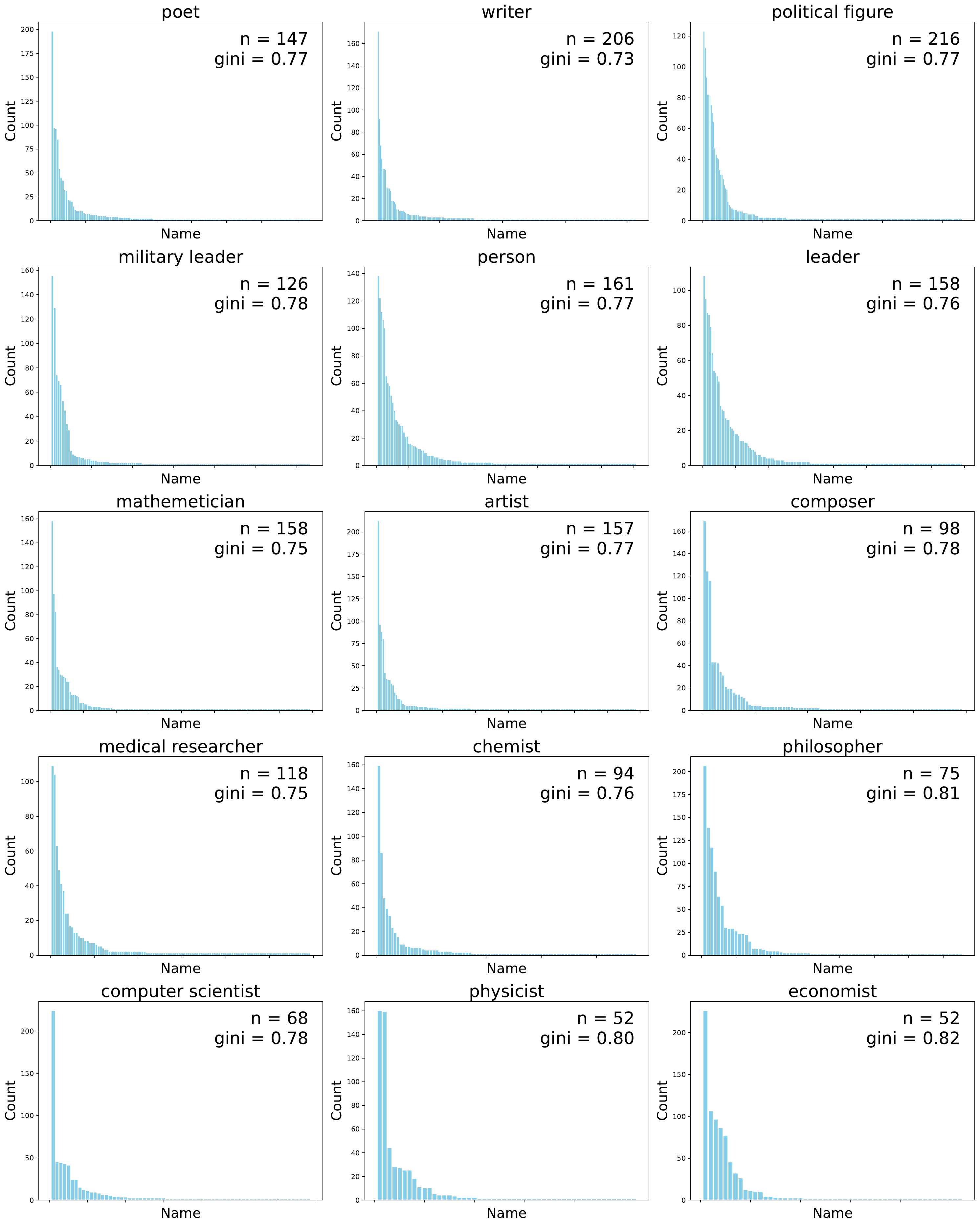}
    \caption{Frequency distribution for every profession (Llama)}
    \label{fig:freq_occ_llama}
\end{figure*}

We visualise the frequency distributions for each LLM separately in Figure~\ref{fig:freq_occ_gpt} (GPT), Figure~\ref{fig:freq_occ_claude} (Claude) and Figure~\ref{fig:freq_occ_llama} (Llama).

Lastly, we visualise the top 10 results for every profession across the different languages, adjectives, LLMs and runs in Table~\ref{tab:results_names}.

\begin{table*}
\centering
\adjustbox{max width=\textwidth}{%
\begin{tabular}{|l|rrr|r|rrrr|r|}
\toprule
 & \multicolumn{4}{c}{Average \#names/response} \vline & \multicolumn{5}{c}{Unique names} \vline \\ \hline
Model & GPT & Claude & Llama & all LLMs & & GPT & Claude & Llama & all LLMs \\ \hline
greatest & 5.24 & 8.72 & 4.63 & 6.20 & & 509& 566 & 613 & 1098\\
most famous & 4.25 & 8.26 & 3.10 & 5.20 & &444 & 492 & 435 & 886 \\
most impactful & 5.11 & 8.48 & 3.45 & 5.68 & &454 & 536 & 505 & 962 \\
most important & 5.34 & 8.81 & 4.13 & 6.09 & &478 & 549 & 503 & 1011\\
most influential & 5.14 & 8.75 & 3.67 & 5.85 & &426 & 539 & 586 & 1023 \\ \hline
All adjectives & 5.01 & 8.60 & 3.80 & 5.80 & Avg. (per adj.) \vline & 462.2 & 536.4 & 528.4 & 998.4\\
 &  &  &  &  & Total (across adj.) \vline &1158 & 1023& 1386  & 2409\\
\bottomrule
\end{tabular}}
\caption{General results by LLM and adjective. We present the average number of names/response and the number of unique names per LLM and adjective. }
\label{tab:general_results_llmadjective}
\end{table*}

\begin{table*}[ht]
\centering
\adjustbox{max width=\textwidth}{%
\begin{tabular}{|l|rrr|r|rrrr|r|}
\toprule
 & \multicolumn{4}{c}{Average \#names/response} \vline & \multicolumn{5}{c}{Unique names} \vline \\ \hline
Language & GPT & Claude & Llama & all LLMs& & GPT & Claude & Llama & all LLMs \\ \hline
Hindi & 5.53 & 7.64 & 2.56 & 5.24 && 369 & 360 & 236 & 618 \\
Spanish & 5.12 & 9.88 & 4.97 & 6.66 & &193 & 293 & 326 & 477\\
Urdu & 5.11 & 6.33 & 3.16 & 4.86 & &419 & 301 & 558 & 918 \\
Russian & 4.85 & 8.47 & 3.04 & 5.45 & &204 & 318 & 272 & 490 \\
English & 4.57 & 9.78 & 9.35 & 7.90 && 165 & 281 & 401 & 520 \\
French & 5.09 & 9.63 & 4.56 & 6.43 & &188 & 309 & 297 & 468 \\
Chinese & 5.75 & 9.75 & 2.30 & 5.93 && 360 & 415 & 224 & 647 \\
Portuguese & 4.22 & 9.71 & 4.45 & 6.13 & &160 & 297 & 410 & 551 \\
Bengali & 5.58 & 8.38 & 2.15 & 5.37 & &371 & 354 & 244 & 642 \\
Arabic & 4.32 & 6.49 & 1.42 & 4.08 & &333 & 289 & 227 & 591 \\ \hline
All languages & 5.02 & 8.60 & 3.80 & 5.80 & Avg. (per lang.) \vline &276.2 & 321.7 & 319.5 &  592.2\\
 &  &  &  &  & Total (across lang.) \vline &1158 & 1023& 1386  & 2409\\
\bottomrule
\end{tabular}}
\caption{General results by LLM and language}
\label{tab:general_results_llmlanguage}
\end{table*}

\begin{table*}[ht]
\centering
\adjustbox{max width=\textwidth}{%
\begin{tabular}{|l|rrr|r|rrrr|r|}
\toprule
 & \multicolumn{4}{c}{Average \#names/response} \vline & \multicolumn{5}{c}{Unique names} \vline \\ \hline
Profession & GPT & Claude & Llama & all LLMs & & GPT & Claude & Llama & all LLMs \\ \hline
Artist & 4.56 & 8.44 & 3.94 & 5.64 && 86 & 96 & 157 & 227 \\
Computer Scientist & 3.50 & 8.06 & 2.39 & 4.65 && 74 & 36 & 68 & 119 \\
Chemist & 3.38 & 6.22 & 2.37 & 3.99 & &83 & 62 & 94 & 174 \\
Composer & 4.56 & 8.11 & 3.47 & 5.38 & &123 & 62 & 98 & 208 \\
Poet & 5.36 & 9.35 & 4.20 & 6.30 & &162 & 225 & 147 & 384 \\
Leader & 7.41 & 8.53 & 5.32 & 7.08 & &120 & 116 & 158 & 260 \\
Physicist & 2.86 & 7.66 & 2.31 & 4.28 & &27 & 53 & 52 & 97 \\
Medical Researcher & 4.42 & 7.29 & 2.85 & 4.85 & &106 & 59 & 118 & 203 \\
Philosopher & 5.97 & 10.50 & 3.90 & 6.79 & &84 & 74 & 75 & 152 \\
Person & 7.13 & 11.52 & 5.99 & 8.21 & &123 & 126 & 161 & 266 \\
Political Figure & 7.72 & 8.34 & 5.92 & 7.33 & &155 & 91 & 216 & 333 \\
Economist & 3.26 & 7.80 & 3.16 & 4.74 & &39 & 31 & 52 & 85 \\
Writer & 5.20 & 9.66 & 4.30 & 6.39 & &190 & 134 & 206 & 365 \\
Military Leader & 5.08 & 8.77 & 3.47 & 5.77 & &109 & 162 & 126 & 286 \\
Mathematician & 4.82 & 8.82 & 3.34 & 5.66 & &90 & 70 & 158 & 237 \\ \hline
All professions & 5.02 & 8.60 & 3.80 & 5.80 & Avg. (per prof.) \vline & 104.7 & 93.1 & 125.7 &  226.4\\
 &  &  &  &  & Total (across lang.) \vline &1158 & 1023& 1386  & 2409\\
\bottomrule
\end{tabular}}
\caption{General results by LLM and profession}
\label{tab:general_results_llmprofession}
\end{table*}

\begin{table*}[ht]
\small
\centering
\caption{Results for each profession. We display the count ($n$) which is the number of responses in which they occur and the percentage (\%) which is the percentage of responses in which they occur ($n/750$).}
\label{tab:results_names}
\begin{adjustbox}{width=\textwidth}
\begin{tabular}{ccc}
\begin{subtable}[t]{0.3\textwidth}
    \centering
    \caption{Artist}
    \begin{tabular}{lrr}
    \toprule
    Name & $n$ & \% \\
    \midrule
    Leonardo Da Vinci & 699 & 93.2 \\
    Pablo Picasso & 508 & 67.7 \\
    Michelangelo & 470 & 62.7 \\
    Vincent Van Gogh & 464 & 61.9 \\
    Rembrandt & 232 & 30.9 \\
    Charles David & 204 & 27.2 \\
    Claude Monet & 171 & 22.8 \\
    Salvador Dali & 112 & 14.9 \\
    William Shakespeare & 77 & 10.3 \\
    Ludwig Beethoven & 76 & 10.1 \\
    \bottomrule
    \end{tabular}
\end{subtable} &
\begin{subtable}[t]{0.3\textwidth}
    \centering
    \caption{Computer Scientist}
    \begin{tabular}{lrr}
    \toprule
    Name & $n$ & \% \\
    \midrule
    Alan Turing & 723 & 96.4 \\
    John Neumann & 364 & 48.5 \\
    Tim Berners & 314 & 41.9 \\
    Lee & 313 & 41.7 \\
    Hopper & 257 & 34.3 \\
    Ada Lovelace & 230 & 30.7 \\
    Dennis Ritchie & 193 & 25.7 \\
    Claude Shannon & 166 & 22.1 \\
    Charles Babbage & 146 & 19.5 \\
    Donald Knuth & 134 & 17.9 \\
    \bottomrule
    \end{tabular}
\end{subtable} &
\begin{subtable}[t]{0.3\textwidth}
    \centering
    \caption{Chemist}
    \begin{tabular}{lrr}
    \toprule
    Name & $n$ & \% \\
    \midrule
    Marie Curie & 509 & 67.9 \\
    Dmitri Mendeleev & 492 & 65.6 \\
    Lavoisier & 474 & 63.2 \\
    Linus Pauling & 274 & 36.5 \\
    Dalton & 180 & 24.0 \\
    Alfred Nobel & 126 & 16.8 \\
    Robert Boyle & 96 & 12.8 \\
    Louis Pasteur & 72 & 9.6 \\
    Frederick Sanger & 54 & 7.2 \\
    Rosalind Franklin & 53 & 7.1 \\
    \bottomrule
    \end{tabular}
\end{subtable} \\

\begin{subtable}[t]{0.3\textwidth}
    \centering
    \caption{Composer}
    \begin{tabular}{lrr}
    \toprule
    Name & $n$ & \% \\
    \midrule
    Mozart & 649 & 86.5 \\
    Beethoven & 617 & 82.3 \\
    Bach & 584 & 77.9 \\
    Wagner & 249 & 33.2 \\
    Chopin & 218 & 29.1 \\
    Tchaikovsky & 204 & 27.2 \\
    Schubert & 179 & 23.9 \\
    Stravinsky & 176 & 23.5 \\
    Debussy & 164 & 21.9 \\
    Brahms & 130 & 17.3 \\
    \bottomrule
    \end{tabular}
\end{subtable} &
\begin{subtable}[t]{0.3\textwidth}
    \centering
    \caption{Poet}
    \begin{tabular}{lrr}
    \toprule
    Name & $n$ & \% \\
    \midrule
    Shakespeare & 565 & 75.3 \\
    Homer & 565 & 75.3 \\
    Dante & 443 & 59.1 \\
    Neruda & 236 & 31.5 \\
    Tagore & 230 & 30.7 \\
    Rumi & 197 & 26.3 \\
    Goethe & 181 & 24.1 \\
    Li Bai & 144 & 19.2 \\
    Virgil & 136 & 18.1 \\
    Whitman & 128 & 17.1 \\
    \bottomrule
    \end{tabular}
\end{subtable} &
\begin{subtable}[t]{0.3\textwidth}
    \centering
    \caption{Leader}
    \begin{tabular}{lrr}
    \toprule
    Name & $n$ & \% \\
    \midrule
    Gandhi & 525 & 70.0 \\
    Mandela & 458 & 61.1 \\
    Churchill & 360 & 48.0 \\
    Lincoln & 355 & 47.3 \\
    Alexander the Great & 347 & 46.3 \\
    Napoleon & 296 & 39.5 \\
    MLK Jr. & 295 & 39.3 \\
    Julius Caesar & 293 & 39.1 \\
    Mao Zedong & 198 & 26.4 \\
    Genghis Khan & 193 & 25.7 \\
    \bottomrule
    \end{tabular}
\end{subtable} \\
\begin{subtable}[t]{0.3\textwidth}
    \centering
    \caption{Physicist}
    \begin{tabular}{lrr}
    \toprule
    Name & $n$ & \% \\
    \midrule
    Einstein & 655 & 87.3 \\
    Newton & 557 & 74.3 \\
    Galileo & 272 & 36.3 \\
    Niels Bohr & 243 & 32.4 \\
    Maxwell & 227 & 30.3 \\
    Feynman & 217 & 28.9 \\
    Hawking & 160 & 21.3 \\
    Marie Curie & 145 & 19.3 \\
    Max Planck & 127 & 16.9 \\
    Faraday & 121 & 16.1 \\
    \bottomrule
    \end{tabular}
\end{subtable} &
\begin{subtable}[t]{0.3\textwidth}
    \centering
    \caption{Medical Researcher}
    \begin{tabular}{lrr}
    \toprule
    Name & $n$ & \% \\
    \midrule
    Louis Pasteur & 571 & 76.1 \\
    Alexander Fleming & 552 & 73.6 \\
    Edward Jenner & 373 & 49.7 \\
    Hippocrates & 295 & 39.3 \\
    Jonas Salk & 270 & 36.0 \\
    Robert Koch & 202 & 26.9 \\
    Leon Harvey & 152 & 20.3 \\
    Marie Curie & 117 & 15.6 \\
    Albert Sabin & 108 & 14.4 \\
    Francis Crick & 99 & 13.2 \\
    \bottomrule
    \end{tabular}
\end{subtable} &
\begin{subtable}[t]{0.3\textwidth}
    \centering
    \caption{Philosopher}
    \begin{tabular}{lrr}
    \toprule
    Name & $n$ & \% \\
    \midrule
    Plato & 695 & 92.7 \\
    Aristotle & 634 & 84.5 \\
    Socrates & 522 & 69.6 \\
    Kant & 516 & 68.8 \\
    Nietzsche & 440 & 58.7 \\
    Descartes & 373 & 49.7 \\
    Confucius & 196 & 26.1 \\
    Sartre & 159 & 21.2 \\
    Karl Marx & 158 & 21.1 \\
    Hegel & 153 & 20.4 \\
    \bottomrule
    \end{tabular}
\end{subtable} \\

\begin{subtable}[t]{0.3\textwidth}
    \centering
    \caption{Person}
    \begin{tabular}{lrr}
    \toprule
    Name & $n$ & \% \\
    \midrule
    Einstein & 556 & 74.1 \\
    Jesus Christ & 531 & 70.8 \\
    Newton & 432 & 57.6 \\
    Muhammad & 422 & 56.3 \\
    Buddha & 353 & 47.1 \\
    Gandhi & 317 & 42.3 \\
    Alexander the Great & 242 & 32.3 \\
    Mandela & 218 & 29.1 \\
    Confucius & 213 & 28.4 \\
    Darwin & 195 & 26.0 \\
    \bottomrule
    \end{tabular}
\end{subtable} &
\begin{subtable}[t]{0.3\textwidth}
    \centering
    \caption{Political Figure}
    \begin{tabular}{lrr}
    \toprule
    Name & $n$ & \% \\
    \midrule
    Gandhi & 534 & 71.2 \\
    Mandela & 464 & 61.9 \\
    Churchill & 425 & 56.7 \\
    Lincoln & 392 & 52.3 \\
    Mao Zedong & 347 & 46.3 \\
    Julius Caesar & 341 & 45.5 \\
    Napoleon & 316 & 42.1 \\
    Alexander the Great & 235 & 31.3 \\
    Hitler & 227 & 30.3 \\
    Lenin & 220 & 29.3 \\
    \bottomrule
    \end{tabular}
\end{subtable} &
\begin{subtable}[t]{0.3\textwidth}
    \centering
    \caption{Economist}
    \begin{tabular}{lrr}
    \toprule
    Name & $n$ & \% \\
    \midrule
    Adam Smith & 722 & 96.3 \\
    Keynes & 519 & 69.2 \\
    Karl Marx & 469 & 62.5 \\
    Friedman & 455 & 60.7 \\
    Ricardo & 262 & 34.9 \\
    Samuelson & 233 & 31.1 \\
    Marshall & 181 & 24.1 \\
    Hayek & 179 & 23.9 \\
    Schumpeter & 171 & 22.8 \\
    Amartya Sen & 96 & 12.8 \\
    \bottomrule
    \end{tabular}
\end{subtable} \\

\end{tabular}
\end{adjustbox}
\end{table*}

\clearpage 

\begin{table*}[ht]
\small
\centering
\ContinuedFloat
\caption{Results for each profession (continued)}
\begin{adjustbox}{width=\textwidth}
\begin{tabular}{ccccc}

\begin{subtable}[t]{0.3\textwidth}
    \centering
    \caption{Writer}
    \begin{tabular}{lrr}
    \toprule
    Name & $n$ & \% \\
    \midrule
    Shakespeare & 659 & 87.9 \\
    Tolstoy & 417 & 55.6 \\
    Homer & 324 & 43.2 \\
    Cervantes & 278 & 37.1 \\
    Dante & 275 & 36.7 \\
    Dickens & 242 & 32.3 \\
    Dostoevsky & 192 & 25.6 \\
    Goethe & 175 & 23.3 \\
    Marquez & 157 & 20.9 \\
    Victor Hugo & 147 & 19.6 \\
    \bottomrule
    \end{tabular}
\end{subtable} &
\begin{subtable}[t]{0.3\textwidth}
    \centering
    \caption{Military Leader}
    \begin{tabular}{lrr}
    \toprule
    Name & $n$ & \% \\
    \midrule
    Alexander the Great & 636 & 84.8 \\
    Napoleon & 596 & 79.5 \\
    Genghis Khan & 531 & 70.8 \\
    Julius Caesar & 506 & 67.5 \\
    Hannibal & 192 & 25.6 \\
    Erwin Rommel & 173 & 23.1 \\
    Sun Tzu & 163 & 21.7 \\
    George Patton & 117 & 15.6 \\
    Saladin & 98 & 13.1 \\
    George Washington & 83 & 11.1\\
    \bottomrule
    \end{tabular}
\end{subtable} &
\begin{subtable}[t]{0.3\textwidth}
    \centering
    \caption{Mathematician}
    \begin{tabular}{lrr}
    \toprule
    Name & $n$ & \% \\
    \midrule
    Newton & 634 & 84.5 \\
    Gauss & 464 & 61.9 \\
    Archimedes & 409 & 54.5 \\
    Euclid & 395 & 52.7 \\
    Euler & 350 & 46.7 \\
    Leibniz & 180 & 24.0 \\
    Einstein & 165 & 22.0 \\
    Hilbert & 159 & 21.2 \\
    Riemann & 146 & 19.5 \\
    Ramanujan & 118 & 15.7 \\
    \bottomrule
    \end{tabular}
\end{subtable} \\


\end{tabular}
\end{adjustbox}
\end{table*}

\end{document}